\documentclass[review,12pt]{elsarticle}
\usepackage{amsmath}
 
\pdfoutput=1
\usepackage{amssymb}
 \usepackage{graphicx}
 \usepackage{algorithm}
 \usepackage{algorithmic}
 \usepackage{xcolor}
 \usepackage{makecell}
 \usepackage{longtable}
 \usepackage{multirow}
 \usepackage{newfloat}
 \usepackage{listings}
 \usepackage{amssymb}
 \usepackage{amsmath}
 \usepackage{makecell}
 \usepackage{xcolor}
 \usepackage {subcaption}
 \usepackage{mathrsfs}  
 \newtheorem{mydef}{Definition}
 \newtheorem{myproof}{proof}

\usepackage{amssymb}
\journal{Pattern Recognition}

\begin{document}

\begin{frontmatter}



\title{Multi-order Graph Clustering with Adaptive Node-level Weight Learning}


\author[1]{Ye Liu \corref{cor1}}
\ead{yliu03@scut.edu.cn}

% \affiliation{organization={School of Future Technology, South China University of Technology},%Department and Organization
%             city={Guangzhou},
%             country={China}}

\author[2]{Xuelei Lin}
\ead{linxuelei@hit.edu.cn}

% \affiliation{organization={School of Science, Harbin Institute of Technology},%Department and Organization
%             city={Shenzhen},
%             country={China}}

\author[1]{Yejia Chen}
\ead{202221062461@mail.scut.edu.cn}

% \affiliation{organization={School of Future Technology, South China University of Technology},%Department and Organization
%             city={Guangzhou},
%             country={China}}

\author[3]{Reynold Cheng}
\ead{ckcheng@cs.hku.hk}

% \affiliation{organization={Department of Computer Science, The University of Hong Kong},%Department and Organization
%             city={Hong Kong},
%             country={China}}

\address[1]{School of Future Technology, South China University of Technology, Guangzhou, China}
\address[2]{School of Science, Harbin Institute of Technology, Shenzhen, China}
\address[3]{Department of Computer Science, The University of Hong Kong, Hong Kong, China}
\cortext[cor1]{Corresponding author}

\begin{abstract}
%Graph clustering is one of the major tasks in graph mining. Existing graph clustering methods focus on lower-order connectivity patterns of individual nodes and edges, while ignoring the higher-order organization at the level of small network subgraph (motif). Recently, some higher-order graph clustering approaches have been designed by constructing a motif-based hypergraph. However, these approaches often suffer from hypergraph fragmentation issue seriously, which would degrade the clustering performance greatly. Moreover, real-world graphs usually consist of diverse motifs, and one node may be involved in more than one motif. The importance of different kinds of motifs for a node varies from node to node. A key challenge under this setting is how to achieve accurate clustering results by integrating information from multiple motifs at the level of node sample. In this paper, we propose a mixed-order graph clustering model (MOGC) to integrate multiple higher-order structures and lower-order structures at the level of node. In particular, with adaptive weights learning mechanism, MOGC is able to automatically adjust the contributions of different motifs for each node. Our method not only tackles hypergraph fragmentation issue but also achieves accurate clustering results. The MOGC model is efficiently solved by an alternating minimization algorithm, the convergence of which is discussed. Experimental results on seven real-world datasets illustrate the effectiveness and efficiency of our proposed method.
Current graph clustering methods emphasize individual node and edge connections, while ignoring higher-order organization at the level of motif. Recently, higher-order graph clustering approaches have been designed by motif-based hypergraphs. However, these approaches often suffer from hypergraph fragmentation issue seriously, which degrades the clustering performance greatly. Moreover, real-world graphs usually contain diverse motifs, with nodes participating in multiple motifs. A key challenge is how to achieve precise clustering results by integrating information from multiple motifs at the node level. In this paper, we propose a multi-order graph clustering model (MOGC) to integrate multiple higher-order structures and edge connections at node level. MOGC employs an adaptive weight learning mechanism to automatically adjust the contributions of different motifs for each node. This not only tackles hypergraph fragmentation issue but enhances clustering accuracy. MOGC is efficiently solved by an alternating minimization algorithm. Experiments on seven real-world datasets illustrate the effectiveness of MOGC. 
\end{abstract}




\end{frontmatter}


\section{Introduction}
    Nowadays, many complex interaction system can be naturally represented as graphs or networks, in which nodes refer to objects to be interested and edges representing the relationship among objects. Graph clustering which is one of the fundamental tasks in graph mining, aims to partition the graph into several densely connected groups with tight internal connections and sparse external connections. Graph clustering has been widely applied in many applications, such as social network, and computer vision.

    Although many graph clustering methods have been proposed \cite{von2007tutorial}, most of them focus on the lower-order structure at the level of individual nodes and edges, and ignore the higher-order structure in the network. An important higher-order structure is network motif, which is defined as the building block of networks. Recently, some motif-based higher-order graph clustering methods  \cite{huang2018harmonic}\cite{tsourakakis2017scalable}\cite{benson2016higher} have been proposed, they often perform clustering on a motif-based hypergraph constructed by utilizing the co-occurence information of two nodes in one motif instance. Motif characterizes higher-order network structures to provide new insights into the organization of complex systems beyond the clustering of nodes based on edges \cite{benson2016higher}. However, previous higher-order clustering methods assume that the motif-based hypergraph is a connected graph, which is not hold in some real-world networks. The motif-based hypergraph usually become fragmented, in which the original connected graph may be fragmented into a large number of connected components and isolated nodes. Then in the process of clustering nodes, these isolated nodes will not be supported by original network, which makes the class label of isolated nodes present randomness. Only a few works \cite{li2019edmot}\cite{li2020community} aim to address the challenges of hypergraph fragmentations, these works all focus on one kind of motif and utilize edges to solve the fragmentation issue. 

\begin{figure*}[ht]
		\vspace{-3cm}
		\centering
		\includegraphics[width=12cm,height=7cm]{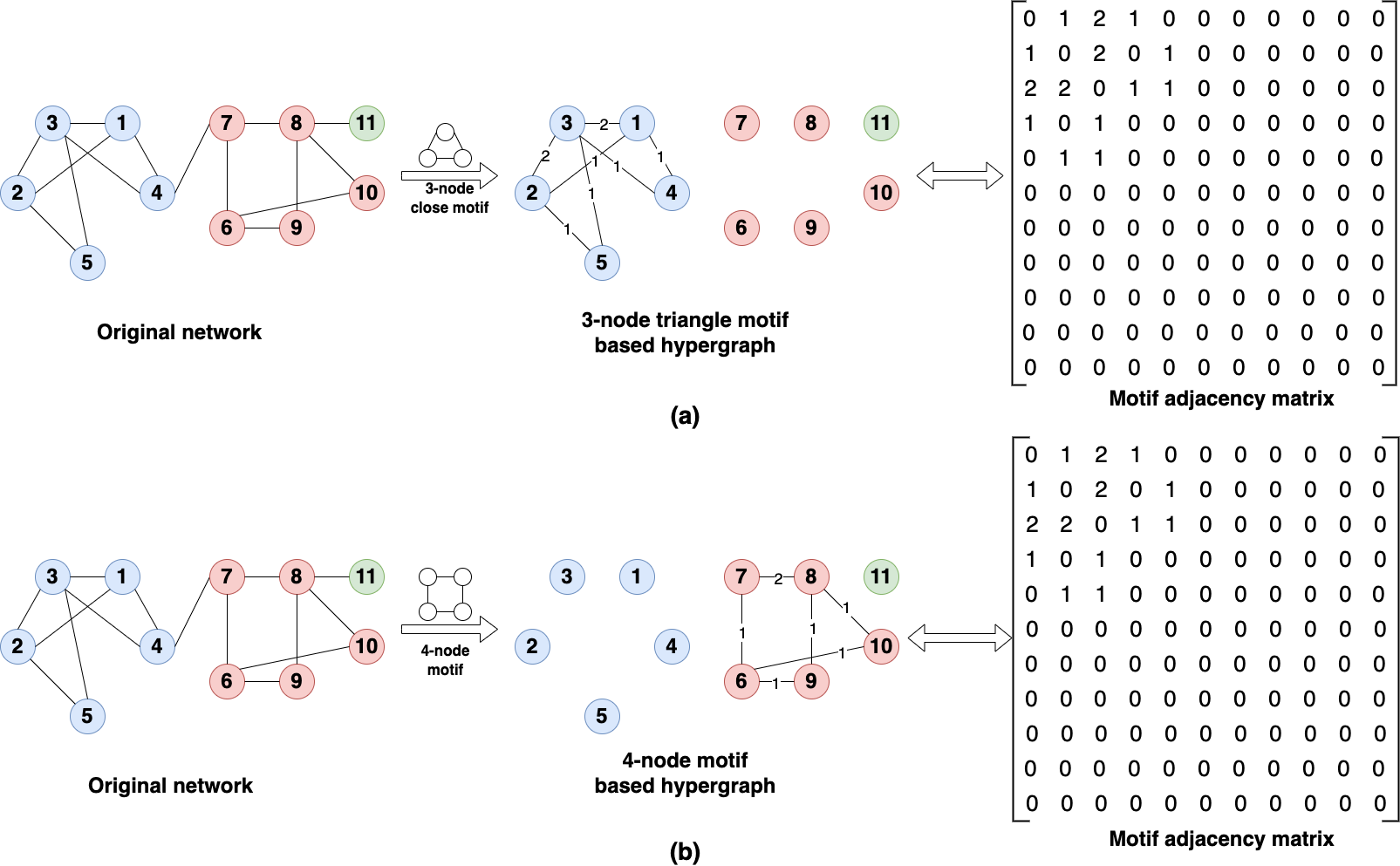}
		\caption{Illustration of fragmentation issue and different motif's contributions to the graph clustering task: (a) Original network is fragmented into one connected component (blue nodes) and six isolated nodes (red nodes and green node) based on 3-node motif. (b) Original network is fragmented into one connected component (red nodes) and six isolated nodes (blue nodes and green node) based on 4-node motif. Original network contains edges with weight 1, numbers on edges demonstrate the edge weight in 3-node motif based hypergraph and 4-node motif based hypergraph, the corresponding motif adjacency matrix is shown in the third column.}\label{motif_example}
		\vspace{-0.5cm}
	\end{figure*}

    However, a large real-world network often consists of many motifs with various functional roles, and different organizational patterns (clustering results) would be revealed with different types of motifs. For example, some 3-node motifs and 4-node motifs are found to play different funcitonal roles in biological systems \cite{benson2016higher}. Moreover, a node is likely to involved in more than one type of network motif, and the importance of different types of motifs for a node varies from node to node in real world \cite{milo2002network}. Therefore, different motifs and sample nodes potentially offer unique contribution to the graph clustering task. For example, in Fig. \ref{motif_example}, blue nodes are closely connected with larger weights based on triangle motif, while another six nodes are isolated. Based on 4-node motif, red nodes are well connected with larger weights, while others are isolated from red nodes. It is obvious that triangle motif based hypergraph could provide more connectivity information for blue nodes, 4-node motif based hypergraphs are useful to identify the cluster of red nodes, different motifs contribute differently for each node in the identification of communities. It should be noted that these differences could be node specific. For example, the isolated node \#11 becomes outlier in triangle motif based hypergraph and 4-node motif based hypergraph, edge-based graph is more informative to the partition of node \#11. 
    For the convenience of description, original edge-based graph is regarded as a kind of motif-based hypergraph in this paper. Thus, in order to achieve accurate clustering result, a more effective approach is to establish a flexible framework in which a weight for each motif of each node can be assigned and adjusted adaptively in the clustering process.
	
    To tackle the above mentioned problems, we propose a framework to integrate multiple higher-order structures (motifs) and lower-order structure with a independent weighting scaler assigned to each motif for each node. The formulated optimization model is composed of two parts, the first term exploits the underlying cluster structure from the integrated hypergraph by evaluating the contribution of each motif to each vertex. Another term is regularization term of motif weight  to each vertex, such that irrelevant hypergraph would be removed in the clustering process. The proposed optimization problem can be solved by alternating scheme, it is decomposed into two subproblems which can be analytically solved respectively. Extensive experiments are conducted on seven datasets, comparison results with existing methods illustrate the effectiveness of the proposed method. The code of MOGC is released in the link\footnote{https://github.com/SCUTFT-ML/MOGC}.

\section{Related Work}
	\subsection{Lower-order Graph Clustering}
	Lower-order graph clustering finds the partition of vertices by utilizing the lower-order connectivity patterns of the graph at the level of nodes and edges. For undirected graph clustering, widely used methods concentrate on spectral clustering \cite{von2007tutorial}, random walk \cite{pons2005computing}, matrix factorization \cite{cai2010graph}, label propagation \cite{wang2007label}, affinity propagation \cite{frey2007clustering}, louvain method \cite{blondel2008fast} and deep learning based methods \cite{cao2016deep}. For example, Nonnegative matrix factorization (NMF) \cite{cai2010graph} factorizes the node adjacency matrix into a basis matrix and a indicator matrix, indicator matrix is the data representation with clustering information. Affinity propogation \cite{frey2007clustering} seeks cluster of nodes by similarities between pairs of nodes.  
	The Louvain method \cite{blondel2008fast} is a greedy optimization of modularity, which measures the relative density of edges inside communities with respect to edges outside communities. Spectral clustering \cite{von2007tutorial} utilizes the eigenvector of graph laplacian matrix to calculate the structure pattern, which is based on graph cut criteria, inculding ratio cut \cite{hagen1992new} and normalized cut \cite{shi2000normalized}. Thus, the main focus of this paper is graph cut criteria based method.
%	Given $\mathbb{G}$ be an undirected graph where $V$ is a set of $N$ vertices, and $E$ represents the set of edges among vertices. In a weighted graph, the corresponding weighted adjacency matrix $\mathbf{A}$ denotes the similarity between each pair of nodes, in which $\mathbf{A}_{i,j}$ denotes the similarity between vertice $V_i$ and vertice $V_j$. Note that $\mathbf{A}_{i,j}=0$ when there is no edge between vertice $V_i$ and $V_j$. The degree of vertice $V_i$ is defined as $d_i=\sum_{j=1}^{N}A_{i,j}$, the degree matrix $\mathbf{D}$ is a diagonal matrix with the degrees $d_1,d_2,\ldots,d_N$ on the main diagonal. The laplacian matrix $\mathbf{L}$ is constructed as $\mathbf{L} = \mathbf{D}-\mathbf{A}$. We consider a $N$-by-$K$ matrix $\mathbf{U}$ as object-cluster-indicator matrix, when the vertice $V_i$ belongs to $k$-th cluster, $\mathbf{U}_{i,j}$ is equal to 1, otherwise $\mathbf{U}_{i,j}$ is equal to 0.  The spectral clustering solve the following optimization problem:
	
%	\begin{equation}\label{spectralclustering}
%		\min_{\mathbf{U}} tr(\mathbf{U}^T\mathbf{L}\mathbf{U})~~~s.t.~~\mathbf{U}^T\mathbf{U}=\mathbf{I}_K
%	\end{equation}
%	Where $tr(.)$ is the trace of a matrix. The optimal solution of Eq.\ref{spectralclustering} is given by the eigenvector corresponding to the first  $K$ smallest eigenvalue of $\mathbf{L}$.

%	The normalized cut aims at solving the following optimization:
%	\begin{equation}\label{normalizedSC}
%		 \mathbf{L}\mathbf{U} = \lambda\mathbf{D}\mathbf{U}
%	\end{equation}
\begin{table}[h]
% 		\vspace{-3cm}
		\centering
		\small{
		 {	\begin{tabular}{c|c}
				\hline
				Notation& Description  \\ \hline
				$\mathcal{G};\mathcal{V};\mathcal{E}  $ & Undirected graph;vertice set; edge set\\
				\hline
				$n;m$;$K$& The number of nodes; the number of motifs;the number of clusters \\ \hline
				$\mathbf{A};\mathbf{D};\mathbf{L} $& Adjacency matrix;degree matrix; laplacian matrix of edge based graph \\ \hline
				$\hat{\mathbf{B}}$;$p$&Adjacency matrix of bidirectional links; the number of nodes in motif\\ \hline
				$\mathbf{B};\mathcal{A}$ & Binary matrix; the set of indices of the anchor nodes in motif.\\ \hline
				$M_p^q, M_p^q(\mathbf{B}, \mathcal{A})$& Motif set\\ \hline
				$\mathbf{A}_{M_p^q}; \mathbf{D}_{M_p^q}; \mathbf{L}_{M_p^q}$& Motif adjacency matrix; motif degree matrix; motif laplacian matrix.\\ \hline
				$\mathbf{W}_{M_p^q}$&Indicator matrix of isolated nodes based on motif $M_p^q$. \\ \hline
				$\Lambda$;$\tilde{\Lambda}$&importance weight of motifs to nodes; the eigenvalue matrix \\ \hline 
				$\mathbf{A}_f$&Adjacency matrix with both higher-order and lower-order structures \\ \hline
				$\mathbf{D}_f; \mathbf{L}_f$&Degree matrix; laplacian matrix of $\mathbf{A}_f$\\ \hline
				$\theta_1, \theta_2$; $\alpha$&Lagragian multiplier parameters; trade-off parameter\\ \hline
		\end{tabular}}}
		\caption{Summary of notations, different notations are separated by a semicolon (;).}
		\label{tab:notations}
\end{table}
	\subsection{{Motif-based Graph Clustering}}
 % Motif was first introduced in \cite{milo2002network}, it is defined as small subgraph with significant large frequency than that in randomized graph preserving the same degree of nodes. Motif has been widely applied in many application scenarios, such as social network, neurobiology.As one of the fundamental tasks in motif-based applications, 
 Many motif-based graph clustering methods have been proposed. 
 % identifies cluster structures in complex systems with higher-order graph structures (motifs). 
 For example, a generalized higher-order graph clustering \cite{benson2016higher} is proposed by extending conductance metric in traditional spectral graph theory to motif conductance. Motif-modularity is proposed to define the class of nodes \cite{arenas2008motif}. \cite{tsourakakis2017scalable} develops triangle conductance and generalizes random walk to triangle based higher-order graph. \cite{yin2017local} proposes motif-based approximated personlized PageRank algorithm to perform local graph clustering with higher-order structures. A graph sparsification method based on motif is designed to improve efficiency and quality of graph clustering \cite{zhao2015gsparsify}. A motif correlation clustering technique for overlapping community detection is developed by minimizing the number of clustering errors associated with both edges and motifs \cite{li2017motif}. { Edge based and motif based multiplex networks are constructed for community detection to reduce information loss during the aggregation of multiplex networks \cite{li2023multiplex}. }
	
	All the above motif-based graph clustering methods ignore the fragmentation issues resulting from motif-based hypergraph construction, which leads to clustering accuracy degradation. Several methods have been proposed to address this issue,
 % Recent works show that motif-based clustering often suffers from the hypergraph fragmentation issues, which results in clustering accuracy degradation. In order to solve this issue, 
 a edge enhancement approach is proposed in \cite{li2019edmot} by injecting original graph edges into the high-order graph. A micro-unit modularity is designed by constructing a micro-unit connection network with integrating both lower-order structure and higher-order structure \cite{huang2020mumod}.  A hybrid-order stochastic block model is designed from the perspective of generative model \cite{wu2021hybrid}. Both an asymmetric triangle and edges are considered as clustering measurements to address fragmentation issue \cite{gao2022graph}. \cite{GE2021mosc} introduces the definition of mixed-order cut criteria, based on it, spectral clustering is performed on the mixed-order adjacency matrix generated by random walks and graph lapalacian. {\cite{wu2024motif} firstly construct sub-higher order network  and the corresponding sub-lower order network based on a kind of motif by eliminating isolated nodes, then community detection is achieved by contrastive learning between above two networks. Finally, the isolated nodes are labelled by label propagation on edge-based graph.}
  
 However, current existing defragmentation methods mainly focus on integrating edges and one kind of motif with ignoring other useful motifs. Therefore, we propose a multi-order graph clustering model (MOGC) to integrate multiple motifs and edge information in order to resolve fragmentation issue and achieve more accurate partition results simultaneously.
\section{Preliminaries}
	Given an undirected graph $\mathcal{G}=\{\mathcal{V},\mathcal{E}\}$, where $\mathcal{V}=\{v_1,\ldots,v_n\}$ is a set of $n$ vertices, and $\mathcal{E}$ represents the set of edges among vertices. The corresponding adjacency matrix $\mathbf{A}\in\mathbb{R}^{n\times n}$ denotes the similarity between each pair of nodes, in which $\mathbf{A}_{i,j}$ is the similarity between vertice $v_i$ and vertice $v_j$. $\mathbf{A}_{i,j}=0$ when there is no edge between vertice $v_i$ and $v_j$. The degree of vertice $v_i$ is defined as $d_i=\sum_{j=1}^{n}\mathbf{A}_{i,j}$, the degree matrix $\mathbf{D}$ is a diagonal matrix with the degrees $d_1,d_2,\ldots,d_n$ on the main diagonal. The laplacian matrix $\mathbf{L}$ is constructed as $\mathbf{L} = \mathbf{D}-\mathbf{A}$.
	\subsection{Network Motif}
\begin{figure}[h]
%			\vspace{-3cm}
		\centering
		\includegraphics[width=12cm,height=3cm]{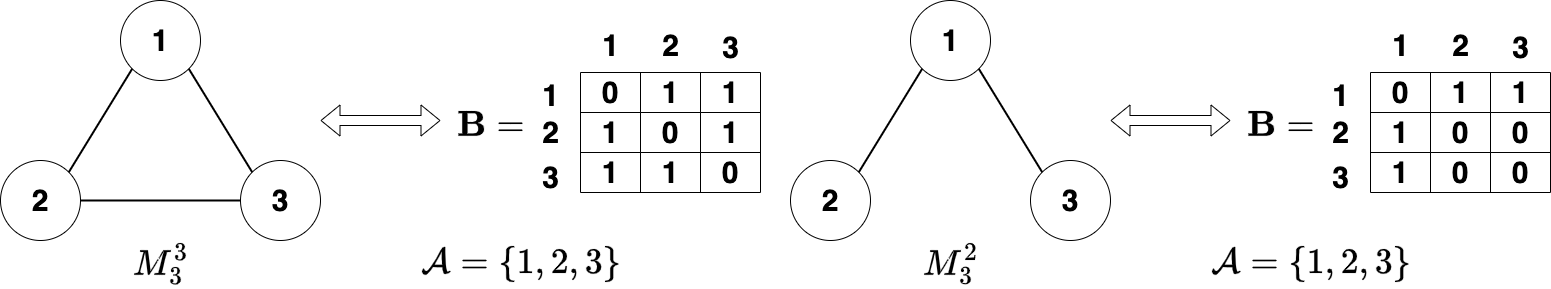}
		\caption{Network motif example for 3-node motifs. ${M}_3^3$ and ${M}_3^2$ are simple motifs, since the anchor set $\mathcal{A}$ contains all of the nodes in ${M}_3^3$ and ${M}_3^2$.  $\mathbf{B}$ is the binary matrix encoding edge pattern of corresponding motif.}\label{motif_def_example}
		\vspace{-0.5cm}
	\end{figure}

	% Different from lower-order structure captured by edges and nodes, 
 The higher-order structure is characterized by motifs, according to \cite{benson2016higher}, a network motif can be defined as:
	\begin{mydef}
		\textbf{Network Motif.} A $p$-node motif $M_p^q$ is defined by a tuple ($\mathbf{B},\mathcal{A}$), $\mathbf{B}$ is a $p\times p$ binary matrix and $\mathcal{A}\subset \{1,2,\ldots,p\}$ { is the set of indices of the anchor nodes}.
	\end{mydef}
	Here $\mathbf{B}$ encodes edge pattern between $p$ nodes, and $\mathcal{A}$ denotes a subset of the $p$ nodes for defining the motif-based adjacency matrix. In other words, two nodes will be regarded as occurring in a given motif only when their indices belong to $\mathcal{A}$. When $\mathcal{A}$ is the entire set of nodes, it is called simple motif, otherwise it is anchored motif. In this paper, we mainly focus on simple motif, a simple example for motif with 3 nodes is shown in Fig. \ref{motif_def_example}.
	\begin{mydef}
		\textbf{Motif Set.} The motif set, denoted as ${M}_p^q(\mathbf{B},\mathcal{A})$, is defined in an undirected graph {$\mathcal{G}$ } as:
		$$
		\begin{array}{l}
			{M}_p^q(\mathbf{B},\mathcal{A})=\{(set(\mathbf{v}),set(\chi_{\mathcal{A}}(\mathbf{v})))|\mathbf{v}\in V^p, {v}_1,\ldots,{v}_p, distinct, \mathbf{A}_{\mathbf{v}}= \mathbf{B}\}
			\end{array}
		$$
	\end{mydef}
	where $\mathbf{v}$ is an vector representing the indices of $p$ nodes, and $\chi_{\mathcal{A}}$ is a selection function taking the subset of a $p$-tuple indexed by $\mathcal{A}$, and set(.) is the operator taking an ordered tuple to an unordered set, e.g., $set((v_1,v_2,\ldots,v_p)=\{v_1,v_2,\ldots,v_p\}$. $\mathbf{A}_{\mathbf{v}}$ is the $p\times p$ adjacency matrix on the subgraph induced by the $p$ nodes of the ordered vector $\mathbf{v}$. 
  %In this paper, we will use $(\mathbf{v},\chi_{\mathcal{A}}(\mathbf{v}))$  to represent $(set(\mathbf{v}),set(\chi_{\mathcal{A}}(\mathbf{v})))$ for simplicity,and only focus on simple motif, i.e., $\chi_{\mathcal{A}}(\mathbf{v})=\mathbf{v}$. 
  Furthermore, any $(set(\mathbf{v}),set(\chi_{\mathcal{A}}(\mathbf{v})))\in {M}_p^q(\mathbf{B},\mathcal{A})$ is a \emph{motif instance}. In this paper, we will use $M_p^q$  to represent $M_p^q(\mathbf{B},\mathcal{A})$ for simplicity.
	 \subsection{Motif Adjacency matrix}
	\begin{mydef}
		\textbf{Motif Adjacency Matrix.} 	Given a motif set {$M_p^q$}, its corresponding motif adjacency matrix is defined as {$(\mathbf{A}_{M_p^q})_{ij}=\sum\limits_{(set(\mathbf{v}),set(\chi_{\mathcal{A}}(\mathbf{v})))}\mathbf{1} ({i,j}\subset set(\chi_{\mathcal{A}}(\mathbf{v}))), \text{for }  i\neq j. $} $1(x)$ is a function, {where} $1(x)$ is 1 when $x$ is true, otherwise it is 0. 
	\end{mydef}
	Note that 
 % weight is added to $(\mathbf{A}_M)_{ij}$ only if nodes $i$ and $j$  appear in the anchor set $\chi_{\mathcal{A}}(\mathbf{v})$, so 
 $(\mathbf{A}_{{M}_p^q})_{ij}$ denotes the number of motif instances when node $v_i$ and $v_j$ co-occur in the same $p$-node motif ${M}_p^q$. The larger $(\mathbf{A}_{{M}_p^q})_{ij}$ is, the more significant the relation between node $v_i$ and $v_j$ is within motif ${M}_p^q$.  A concrete example of motif-based adjacency matrix is shown in Fig. \ref{motif_example}. After the motif-based adjacency matrix is generated, the motif diagonal degree matrix is defined as {$(\mathbf{D}_{{M}_p^q})_{ii}=\sum_{j=1}^{n}(\mathbf{A}_{{M}_p^q})_{ij}$} and the motif Laplacian is {$\mathbf{L}_{{M}_p^q}=\mathbf{D}_{{M}_p^q}-\mathbf{A}_{{M}_p^q}$}.  
	
	Generally, the construction of $\mathbf{A}_{{M}_p^q}$ is related to subgraph counting in large graphs \cite{jha2015path}\cite{wang2016minfer}\cite{kashtan2004efficient}. We explore motifs with different sizes, including 3-node, 4-node and 5-node motifs { in this paper}. 
 % The adjacency matrix of 3-node motifs can be computed based on the matrix computation in \cite{benson2016higher}. For 4-node and 5-node motifs, it is very expensive to enumerate all of the corresponding subgraphs, so we adopt an approximation algorithm to compute 4-node and 5-node motif-based matrices according to a sampling method \cite{kashtan2004efficient}.
	\subsubsection{3-node Motif-based Adjacency Matrix Construction}
 \label{compute}
  % In this paper, we only consider undirected graph, so 
  {For undirected graph $\mathcal{G}$, there are two different kinds of 3-node motifs shown in Fig. \ref{motif_def_example}}. Let $\hat{\mathbf{B}}$ is the adjacency matrix of the bidirectional links of $\mathcal{G}$, {then $\hat{\mathbf{B}}=\mathbf{A}\circ \mathbf{A}^T$ according to \cite{benson2016higher},}
  % According to \cite{benson2016higher}, $\hat{\mathbf{B}}=\mathbf{A}\circ \mathbf{A}^T$,
  where $\circ$ denotes the Hadamard product. For the motif ${M}_3^3$, the corresponding motif-based adjacency matrix $\mathbf{A}_{M_3^3}=(\hat{\mathbf{B}}\cdot\hat{\mathbf{B}})\circ \hat{\mathbf{B}}$.  It is very efficient to generate ${M}_3^3$ motif-based adjacency matrix when the edge based adjacency matrix $\mathbf{A}$ is sparse.  For the motif ${M}_3^2$, we obtain the bidirectional links of  $\mathcal{G}$, then enumerate all bidirectional links to 
%  , i.e., $\hat{\mathbf{B}}$, we can run Algorithm 1 in \cite{zhao2019ranking} to 
  generate ${M}_3^2$ motif-based adjacency matrix $\mathbf{A}_{M_3^2}$ according to \cite{benson2016higher}.

\subsubsection{4-node and 5-node Motif-based Adjacency Matrix Construction}
{ For a motif with $p$ nodes, $p=4$ or $p=5$, we need count all motif instances in the graph to calculate its corresponding motif-based adjacency matrix. }
%In order to calculate {\color{blue}4-node and 5-node} motif-based adjacency matrix, we need count all motif instances in the graph. Naively, for a motif with $p$ nodes, 
$\mathcal{O}(n^p)$ time is needed for enumerating all motif instances, the computation cost will increase dramatically with the increase of number of nodes.  Moreover, 
%for a motif with a given number of nodes, 
there are too many isomorphic types of motifs. For example, there are  199 4-node motifs, 9364 5-node motifs \cite{kashtan2004efficient}.  Several methods with available softwares have been proposed to discover motif efficiently \cite{jha2015path}\cite{wang2016minfer}\cite{kashtan2004efficient}. {Hence, for the construction of $p$-node motif-based adjacency matrix, we firstly utilize a sampling method \cite{kashtan2004efficient} with software mFinder\footnote[1]{https://www.weizmann.ac.il/mcb/UriAlon/download/network-motif-software} to detect motifs, which samples numerous $p$-node motifs based on the probability distribution derived from the frequency of various types of $p$-node motifs. Besides, mFinder also can generate all motif instances for each type of $p$-node motifs. Finally, we input all motif instances into Algorithm 1 in \cite{zhao2019ranking} to construct $p$-node motif-based adjacency matrix.}

     \section{The Proposed Method}
	%Higher-order clustering methods usually suffer from hypergraph fragmentation issue. Thus, the importance of different motifs for different nodes in graph clustering is different, so we propose a graph clustering method with node-level adaptive motif weights to effectively integrate multiple higher-order structure and lower-order structure, such that hypergraph fragmentation issue could be tracled and clustering performance could be further improved. 
	%
	%\subsection{}
	Given a graph $\mathcal{G}=\{\mathcal{V},\mathcal{E}\}$ and a motif ${M}_p^q$, we can obtain a symmetric and weighted motif adjacency matrix $\mathbf{A}_{{M}_p^q}\in \mathbb{R}^{n\times n}$. Note that motif ${M}_p^q$ can be higher-order structure or lower-order structure, e.g., edge, 3-node motif, 4-node motif.  In order to tackle the hypergraph fragmentation issue, we introduce a diagonal matrix $\mathbf{W}_{{M}_p^q}$ to denote the isolated nodes based on motif ${M}_p^q$, if the $i$-th node is a isolated node in the {corresponding} higher-order graph, then $(\mathbf{W}_{{M}_p^q})_{i,i}=0$, otherwise $(\mathbf{W}_{{M}_p^q})_{i,i}=1$. 
	
	Moreover, assuming there are $m$ different motifs ${M}_{p_j}^{q_j}$ with $j\in[1,m]$, 
 % since the contribution of different motifs for a node varies from node to node, 
 we introduce a nonnegative matrix $\mathbf{\Lambda}\in\mathbb{R}^{n\times m}$ to denote the weight of each motif for each node. Each element $\mathbf{\Lambda}_{i,j}$ ( $\forall i\in [1,n], j\in[1,m], \mathbf{\Lambda}_{i,j}\geq 0$) is the weight of motif $\mathbf{M}_{p_j}^{q_j}$ for node $v_i$, {which illustrates the importance of motif $\mathbf{M}_{p_j}^{q_j}$ to node $v_i$}. For example, in Fig. \ref{motif_example}, the node \#11  has connectivity with other nodes in edge-based graph, while it becomes outlier in 3-node motif based hypergraph and 4-node motif based hypergraph. Therefore, the importance of edge, 3-node motif, and 4-node motif for node \#11 are different, and the weight of edge for node \#11 should be highest.
% Besides, we normalize all the weights of motifs for each node such that $\sum_{j=1}^M\mathbf{\Lambda}_{i,j}=1,\forall i \in [1,N]$. 
Note that if $i$-th node is a isolated node under motif ${M}_{p_j}^{q_j}$, then we set $\mathbf{\Lambda}_{i,j}=0$. Therefore, to retain both higher-order structure and lower-order structure, we fuse the information from $m$ motifs by $\sum_{j=1}^m\mathbf{W}_{{M}_{p_j}^{q_j}}\mathbf{A}_{{M}_{p_j}^{q_j}}diag(\mathbf{\Lambda}_{:,j})$. In order to guarantee symmetry of combined adjacent matrix, we set:
	\begin{equation}\label{A}
		{\mathbf{A}_f}=\frac{1}{2}\sum_{j=1}^m(\mathbf{W}_{{M}_{p_j}^{q_j}}\mathbf{A}_{{M}_{p_j}^{q_j}}diag(\mathbf{\Lambda}_{:,j})+diag(\mathbf{\Lambda}_{:,j})\mathbf{A}_{{M}_{p_j}^{q_j}}\mathbf{W}_{{M}_{p_j}^{q_j}})
	\end{equation}
	Here, $diag()$ converts a vector into an diagonal matrix with vector as the diagonal elements, and $\mathbf{\Lambda}_{:,j}$ is the $j$-th column of $\mathbf{\Lambda}$, it denotes the importance weight of motif ${M}_{p_j}^{q_j}$ for every node in the graph.  After obtaining fushed symmetric adjacency matrix ${\mathbf{A}_f}$, we can apply popular spectral clustering on fused adjacency matrix, it is formulated as:
	\begin{eqnarray}\label{opt-1}
		\min_{\mathbf{\Lambda},\mathbf{U}}tr({\mathbf{U}^T{\mathbf{L}_f}\mathbf{U}}) \quad
		s.t. \quad \mathbf{U}^T\mathbf{D}_f\mathbf{U}=\mathbf{I} 
	\end{eqnarray}
	Here ${\mathbf{L}_f=\mathbf{D}_f-\mathbf{A}_f}$, ${\mathbf{D}_f}$ is a diagonal degree matrix, $tr()$ denotes the trace, $\mathbf{U}\in\mathbb{R}^{n\times K}$ is the indicator matrix, which contains the clustering information. {Eq. \eqref{opt-1}} can avoid the degenerated case that all nodes are partitioned into one connected component \cite{shi2000normalized}.
 % The constrain $\mathbf{U}^T{\color{blue}\mathbf{D}_f}\mathbf{U}=\mathbf{I}$ is designed to avoid the degenerated case that the embedding $\mathbf{U}$ of all nodes are equal, which results in all nodes are partitioned into one connected components.
 The combined adjacent matrix ${\mathbf{A}_f}$ depends on unknown variable $\mathbf{\Lambda}$, which denotes the importance weight of each motif for each node. {To determine appropriate weights that mitigate the influence of irrelevant motifs while reinforcing meaningful ones, we constrain the sum of all motif weights for each node equal to one}, i.e., $\sum_{j=1}^m \mathbf{\Lambda}_{i,j}=1, \forall i\in [1,n]$. Finally, the multi-order graph clustering (MOGC) with adaptive weights learning mechanism is formulated as:
	\begin{equation}\label{opt}
		\min_{\mathbf{\Lambda},\mathbf{U}} tr({\mathbf{U}^T\mathbf{L}_f\mathbf{U}})+\alpha\|\mathbf{\Lambda}\|_F^2, ~~s.t.~~\mathbf{U}^T\mathbf{D}_f\mathbf{U}=\mathbf{I}, ~\mathbf{\Lambda}\mathbf{1}_m=\mathbf{1}_{n},  \mathbf{\Lambda} \geq 0
	\end{equation}
	 $\mathbf{1}_m$ is a vector of length $m$ with all the elements being 1, $\mathbf{I}$ is the identity matrix, the second term $\|\mathbf{\Lambda}\|_F^2$ forces a smooth solution, such that irrelevant hypergraphs would be removed in the clustering process. $\alpha$ is a trade-off parameter. 
	\section{Optimization}
	%Let $\tilde{\mathbf{A}}=[\mathbf{W}^1\mathbf{A}^1, \mathbf{W}^2\mathbf{A}^2,\cdots,\mathbf{W}^M\mathbf{A}^M]^T$, 
	%$\tilde{\mathbf{\Lambda}}=[diag(\mathbf{\Lambda}_{(:,1)}),\ldots,diag(\mathbf{\Lambda}_{(:,M)})]$, according to Eq. \eqref{A}, it is easy to know that  $\mathbf{A}=\frac{1}{2}(\tilde{\mathbf{\Lambda}}\tilde{\mathbf{A}}+\tilde{\mathbf{A}}^T\tilde{\mathbf{\Lambda}}^T)$, 
	It is difficult to solve Eq. \eqref{opt} directly, since both $\mathbf{L}_f$ and $\mathbf{D}_f$ depend on unknown variable $\mathbf{\Lambda}$. We use alternating optimization to solve Eq. \eqref{opt}. 
	\subsection{Fix $\mathbf{\Lambda}$, solve $\mathbf{U}$:}
	When $\mathbf{\Lambda}$ is fixed, the minimization problem Eq.\eqref{opt} would become:
	\begin{equation}\label{opt1}
		\min_{\mathbf{U}}~tr(\mathbf{U}^T\mathbf{L}_f\mathbf{U})~~s.t.~~\mathbf{U}^T\mathbf{D}_f\mathbf{U}=\mathbf{I}.
	\end{equation}
	Since $\mathbf{L}_f$ is symmetric, Eq. \eqref{opt1} is degenerated to the generalized eigenvalue problem \cite{shi2000normalized} as follows:
	\begin{equation}\label{generalizedeigenvalue}
		\mathbf{L}_f\mathbf{U}=\mathbf{D}_f\mathbf{U}\tilde{\mathbf{\Lambda}}
	\end{equation}
	where $\tilde{\mathbf{\Lambda}}=diag(\tilde{\lambda_1},\tilde{\lambda_2},\ldots,\tilde{\lambda_n})$ is the eigenvalue matrix, the solution of $\mathbf{U}$ is given by the corresponding K eigevectors corresponding to the first K smallest eigenvalues of generalized eigenvalue problem Eq. \eqref{generalizedeigenvalue}.
\subsection{ Fix $\mathbf{U}$, solve $\mathbf{\Lambda}$:}
When $\mathbf{U}$ is fixed, the minimization problem Eq. \eqref{opt} would become:
	\begin{equation}\label{opt2} 
		\min_{\mathbf{\Lambda}} tr(\mathbf{U}^T\mathbf{L}_f\mathbf{U})+\alpha\|\mathbf{\Lambda}\|_F^2~~s.t.~~\mathbf{U}^T\mathbf{D}_f\mathbf{U}=\mathbf{I},\mathbf{\Lambda}{\mathbf{1}}_m={\mathbf{1}}_n, \mathbf{\Lambda}\geq 0
	\end{equation}
	Let  $\mathbf{U}_k$ denotes the $k$-th column of $\mathbf{U}$.
%	then Eq. \eqref{opt2} would become
%	\begin{equation}\label{opt3}
%		\min_{\mathbf{\Lambda}}\sum_{k=1}^{K}\mathbf{U}_k^T\mathbf{L}_f\mathbf{U}_k+\alpha\|\mathbf{\Lambda}\|_F^2 ~~s.t.~~\mathbf{U}_k^T\mathbf{D}_f\mathbf{U}_k=1,\mathbf{\Lambda}{\mathbf{1}}_m={\mathbf{1}}_n,\mathbf{\Lambda}\geq 0
%	\end{equation}
	Because $\mathbf{A}_f$ satisfies Eq. \eqref{A}, and $\mathbf{L}_f=\mathbf{D}_f-\mathbf{A}_f=diag(\mathbf{A}_f{\mathbf{1}_n})-\mathbf{A}_f$. Then Eq. \eqref{opt2} would become
{	\begin{eqnarray}\label{opt4} 
		&&\min_{\mathbf{\Lambda}}\sum_{k=1}^{K} \mathbf{U}_k^Tdiag(\mathbf{A}_f{\mathbf{1}_N})\mathbf{U}_k-\sum_{k=1}^{K}\mathbf{U}_k^T\mathbf{A}_f\mathbf{U}_k+\alpha\|\mathbf{\Lambda}\|_F^2 \\ \nonumber &&~s.t.~\mathbf{U}_k^T\mathbf{D}_f\mathbf{U}_k=1,\mathbf{\Lambda}{\mathbf{1}}_m={\mathbf{1}}_n,\mathbf{\Lambda}\geq 0
	\end{eqnarray}}
	Let  $\tilde{\mathbf{A}}_k=[\mathbf{U}_k^T\mathbf{W}_{{M}_{p_1}^{q_1}}\mathbf{A}_{{M}_{p_1}^{q_1}}diag(\mathbf{U}_k),\ldots,\mathbf{U}_k^T\mathbf{W}_{{M}_{p_m}^{q_m}}\mathbf{A}_{{M}_{p_m}^{q_m}}diag(\mathbf{U}_k)]$, and $\tilde{\mathbf{\Lambda}}=[\mathbf{\Lambda}^T_{:,{M}_{p_1}^{q_1}},\ldots,\mathbf{\Lambda}^T_{:,{M}_{p_m}^{q_m}}]^T$, then 
%	For the term $\sum_{i=1}^{K}\mathbf{U}_k^T\mathbf{A}_f\mathbf{U}_k$, it is 
	\begin{eqnarray}\label{opt5}\nonumber
		\sum_{k=1}^{K}\mathbf{U}_k^T\mathbf{A}_f\mathbf{U}_k&=&\frac{1}{2}\sum_{k=1}^{K}\sum_{j=1}^{m}\mathbf{U}_k^T[diag(\mathbf{\Lambda}_{:,{M}_{p_j}^{q_j}})\mathbf{A}_{{M}_{p_j}^{q_j}}\mathbf{W}_{{M}_{p_j}^{q_j}} 
		+\mathbf{W}_{{M}_{p_j}^{q_j}}\mathbf{A}_{{M}_{p_j}^{q_j}}diag(\mathbf{\Lambda}_{:,{M}_{p_j}^{q_j}})]\mathbf{U}_k \\ 
		%	&=&\frac{1}{2}\sum_{i=1}^{K}\sum_{m=1}^{M}\mathbf{\Lambda^T(:,m)}diag(\mathbf{U}_i)\mathbf{A}^{(m)}\mathbf{W}^{(m)}\mathbf{U}_i \\ \nonumber &&+\mathbf{U}_i^T\mathbf{W}^{(m)}\mathbf{A}^{(m)}diag(\mathbf{U}_i)\mathbf{\Lambda(:,m)} \\ \nonumber
		&=&\sum_{k=1}^{K}\sum_{j=1}^m\mathbf{U}_k^T\mathbf{W}_{{M}_{p_j}^{q_j}}\mathbf{A}_{{M}_{p_j}^{q_j}}diag(\mathbf{U}_k)\mathbf{\Lambda}_{:,{M}_{p_j}^{q_j}} = \sum_{k=1}^K\tilde{\mathbf{A}_k}\tilde{\mathbf{\Lambda}}
	\end{eqnarray}
%	 Eq. \eqref{opt5} becomes
%	$$\sum_{k=1}^K\mathbf{U}_k^T\mathbf{A}_f\mathbf{U}_k=\sum_{k=1}^K\tilde{\mathbf{A}_k}\tilde{\mathbf{\Lambda}}
%	$$
	Similarly, it is easy to derive that 
	$diag(\mathbf{A}_f{\mathbf{1}_n})= \frac{1}{2}\sum_{j=1}^{m}diag(\mathbf{\Lambda}_{:,{M}_{p_j}^{q_j}})diag(\mathbf{A}_{{M}_{p_j}^{q_j}}\mathbf{W}_{{M}_{p_j}^{q_j}}{\mathbf{1}_n})+diag(\mathbf{W}_{{M}_{p_j}^{q_j}}\mathbf{A}_{{M}_{p_j}^{q_j}}diag(\mathbf{\Lambda}_{:,{M}_{p_j}^{q_j}}){\mathbf{1}_n})$, then 
	\begin{eqnarray}\label{opt6}\nonumber
		\sum_{k=1}^{K}\mathbf{U}_k^Tdiag(\mathbf{A}_f{\mathbf{1}_n})\mathbf{U}_k
%			&=&\frac{1}{2}\sum_{k=1}^K\sum_{j=1}^M\mathbf{U}_k^Tdiag(\mathbf{\Lambda}_{:,\mathbf{M}_{p_j}^{q_j}})diag(\mathbf{A}_{\mathbf{M}_{p_j}^{q_j}}\mathbf{W}_{\mathbf{M}_{p_j}^{q_j}}{\mathbf{1}_N})\mathbf{U}_k+\mathbf{U}_k^Tdiag(\mathbf{W}_{\mathbf{M}_{p_j}^{q_j}}\mathbf{A}_{\mathbf{M}_{p_j}^{q_j}}diag(\mathbf{\Lambda}_{:,\mathbf{M}_{p_j}^{q_j}}){\mathbf{1}_N})\mathbf{U}_k \\ \nonumber
%			&=&\frac{1}{2}\sum_{k=1}^{K}\sum_{j=1}^{M}\mathbf{U}_k^Tdiag(\mathbf{A}_{\mathbf{M}_{p_j}^{q_j}}\mathbf{W}_{\mathbf{M}_{p_j}^{q_j}}{\mathbf{1}_N})diag(\mathbf{U}_k)\mathbf{\Lambda}_{:,\mathbf{M}_{p_j}^{q_j}}+\mathbf{U}_k^Tdiag(\mathbf{U}_k)\mathbf{W}_{\mathbf{M}_{p_j}^{q_j}}\mathbf{A}_{\mathbf{M}_{p_j}^{q_j}}diag(\mathbf{\Lambda}_{:,\mathbf{M}_{p_j}^{q_j}}){\mathbf{1}_N}\\ \nonumber
		&=&\frac{1}{2}\sum_{k=1}^{K}\sum_{j=1}^{m}\mathbf{U}_k^Tdiag(\mathbf{A}_{{M}_{p_j}^{q_j}}\mathbf{W}_{\mathbf{M}_{p_j}^{q_j}}{\mathbf{1}_n})diag(\mathbf{U}_k)\mathbf{\Lambda}_{:,{M}_{p_j}^{q_j}} \\ 
		&&+\mathbf{U}_k^Tdiag(\mathbf{U}_k)\mathbf{W}_{{M}_{p_j}^{q_j}}\mathbf{A}_{{M}_{p_j}^{q_j}}diag({\mathbf{1}_n})\mathbf{\Lambda}_{:,{M}_{p_j}^{q_j}}
	\end{eqnarray}
	Let $\hat{\mathbf{A}}_k=[\mathbf{U}_k^Tdiag(\mathbf{A}_{{M}_{p_1}^{q_1}}\mathbf{W}_{{M}_{p_1}^{q_1}}{\mathbf{1}_n})diag(\mathbf{U}_k),\cdots,$
	$\mathbf{U}_k^Tdiag(\mathbf{A}_{{M}_{p_m}^{q_m}}\mathbf{W}_{{M}_{p_m}^{q_m}}{\mathbf{1}_n})diag(\mathbf{U}_k)]$, \\
 $\bar{\mathbf{A}}_k=[\mathbf{U}_k^Tdiag(\mathbf{U}_k)\mathbf{W}_{{M}_{p_1}^{q_1}}\mathbf{A}_{{M}_{p_1}^{q_1}},\ldots,$
	$\mathbf{U}_k^Tdiag(\mathbf{U}_k)\mathbf{W}_{{M}_{p_m}^{q_m}}\mathbf{A}_{{M}_{p_m}^{q_m}}]$, then 
%	Eq. \eqref{opt6} becomes 
	$$
	\sum_{k=1}^{K} \mathbf{U}_k^Tdiag(\mathbf{A}_f{\mathbf{1}_N})\mathbf{U}_k
	%=\frac{1}{2}\sum_{k=1}^K\hat{\mathbf{A}_k}\tilde{\mathbf{\Lambda}}+\bar{\mathbf{A}_k}\tilde{\mathbf{\Lambda}}
	=\frac{1}{2}\sum_{k=1}^K(\hat{\mathbf{A}}_k+\bar{\mathbf{A}}_k)\tilde{\mathbf{\Lambda}}
	$$
	Finally, according to Eq. \eqref{opt5} and Eq. \eqref{opt6}, the optimization problem Eq. \eqref{opt4} can be solved instead of 
 \begin{equation}\label{opt7}
%		&&\min_{\tilde{\mathbf{\Lambda}}}(\frac{1}{2} \sum_{i=1}^K(\hat{\mathbf{A}}_i+\bar{\mathbf{A}}_i)-\sum_{i=1}^K\tilde{\mathbf{A}}_i)\tilde{\mathbf{\Lambda}}+\alpha\|\tilde{\mathbf{\Lambda}}\|_F^2\\ \nonumber
%			&s.t.& (\hat{\mathbf{A}_k}+\bar{\mathbf{A}_k})\tilde{\mathbf{\Lambda}}=1, \mathbf{M}\tilde{\mathbf{\Lambda}}={\mathbf{1}_N},  0 \leq \tilde{\mathbf{\Lambda}}, \forall k\in [1,K] \\ \nonumber
		 \min_{\tilde{\mathbf{\Lambda}}}\mathbf{V}\tilde{\mathbf{\Lambda}}+\alpha\|\tilde{\mathbf{\Lambda}}\|_F^2
	~ s.t.~ (\hat{\mathbf{A}}_k+\bar{\mathbf{A}}_k)\tilde{\mathbf{\Lambda}}=1, \mathbf{M}\tilde{\mathbf{\Lambda}}={\mathbf{1}_n},   \tilde{\mathbf{\Lambda}} \geq 0, \forall~ k\in [1,K] 
	\end{equation}
Here $\mathbf{V}=\frac{1}{2}\sum_{i=1}^{K}({\hat{\mathbf{A}}_i}+{\bar{\mathbf{A}}_i})-\sum_{i=1}^K{\tilde{\mathbf{A}}_i}$ with $\mathbf{V}\in \mathbb{R}^{1\times nm}$, and $\mathbf{M}=[\mathbf{I}_n,\ldots,\mathbf{I}_n]$ with $\mathbf{M}\in \mathbb{R}^{n\times nm}$. Let 
	$\mathbf{P}=  \left[ (\hat{\mathbf{A}}_1+\bar{\mathbf{A}}_1)^T, (\hat{\mathbf{A}}_2+\bar{\mathbf{A}}_2)^T,\ldots,  (\hat{\mathbf{A}}_K+\bar{\mathbf{A}}_K)^T
	\right]^T
	$.
%	For the constrain $(\hat{\mathbf{A}_k}+\bar{\mathbf{A}_k})\tilde{\mathbf{\Lambda}}=1, \forall~k \in [1,K]$, we can use the $\mathbf{B}\tilde{\mathbf{\Lambda}}=\mathbf{1}_K$ instead. 
	Then for Eq. \eqref{opt7}, it becomes
	\begin{eqnarray}\label{optoflambda}
		&\min\limits_{\tilde{\mathbf{\Lambda}}}&\mathbf{V}\tilde{\mathbf{\Lambda}}+\alpha\|\tilde{\mathbf{\Lambda}}\|_F^2
		\quad s.t. \quad \mathbf{P}\tilde{\mathbf{\Lambda}}=\mathbf{1}_K, \mathbf{M}\tilde{\mathbf{\Lambda}}={\mathbf{1}_n},   \tilde{\mathbf{\Lambda}}  \geq 0
	\end{eqnarray}
	Denote $\Phi_1\in \mathbb{R}^{n\times 1},\Phi_2\in \mathbb{R}^{K\times 1}$ as the Lagrangian multiplier, then we can write the Lagrangian function of Eq. \eqref{optoflambda} as
	\begin{equation}\label{opt8}
		\mathbb{F}=\mathbf{V}\tilde{\mathbf{\Lambda}}+\alpha\|\tilde{\mathbf{\Lambda}}\|_F^2-\Phi_1^T(\mathbf{M}\tilde{\mathbf{\Lambda}}-{\mathbf{1}_N})-\Phi_2^T(\mathbf{P}\tilde{\mathbf{\Lambda}}-\mathbf{1}_K)-\xi_{+}(\tilde{\mathbf{\Lambda}})
	\end{equation}
	$\xi_{+}$ represents the delta function which provides $+\infty$ to the negative value. The derivative of Eq. \eqref{opt8} with respect to $\tilde{\mathbf{\Lambda}}$ is
	\begin{equation}\label{opt9}
		\frac{\partial \mathbb{F}}{\partial \tilde{\mathbf{\Lambda}}}=2\alpha\tilde{\mathbf{\Lambda}}+\mathbf{V}^T-\mathbf{M}^T\Phi_1-\mathbf{P}^T\Phi_2
	\end{equation}
	With KKT condition,the optimal solution can be defined as
	%\begin{equation}\label{solution}
	%\tilde{\mathbf{\Lambda}}=\frac{\mathbf{M}^T\Phi_1+\mathbf{B}^T\Phi_2-\mathbf{V}^T}{2\alpha}
	%\end{equation}
	%Since $\tilde{\mathbf{\Lambda}} \geq 0$, then 
	\begin{equation}\label{lambdasolution}
		\tilde{\mathbf{\Lambda}}=Proj(\frac{\mathbf{M}^T\Phi_1+\mathbf{P}^T\Phi_2-\mathbf{V}^T}{2\alpha})_{+}
	\end{equation}
	Where $Proj(\tilde{\mathbf{\Lambda}})_{+}$ indicates that if $\tilde{\mathbf{\Lambda}}< 0$, then set $\tilde{\mathbf{\Lambda}}=0$. 
	Since $\mathbf{M}\tilde{\mathbf{\Lambda}}={\mathbf{1}}_{n},\mathbf{P}\tilde{\mathbf{\Lambda}}=\mathbf{1}_K$, then it is easy to derive that 
%	$
%	\mathbf{M}(\frac{\mathbf{M}^T\Phi_1+\mathbf{B}^T\Phi_2-\mathbf{V}^T}{2\alpha})={\mathbf{1}}_{N}
%	$
%	and
%	$
%	\mathbf{B}(\frac{\mathbf{M}^T\Phi_1+\mathbf{B}^T\Phi_2-\mathbf{V}^T}{2\alpha})={\mathbf{1}}_{K}
%	$, then 
	the optimal $\Phi_1$ and $\Phi_2$ are
	\small{
	\begin{eqnarray}\label{phi} \nonumber
\Phi_1&=&[\mathbf{M}\mathbf{M}^T-\mathbf{M}\mathbf{P}^T(\mathbf{P}\mathbf{P}^T)^{-1}\mathbf{P}\mathbf{M}^T]^{-1}	[2\alpha\mathbf{1}_n+\mathbf{M}\mathbf{V}^T-\mathbf{M}\mathbf{P}^T(\mathbf{P}\mathbf{P}^T)^{-1}(2\alpha\mathbf{1}_K+\mathbf{P}\mathbf{V}^T)] \\ \nonumber
\Phi_2&=&[\mathbf{P}\mathbf{P}^T-\mathbf{P}\mathbf{M}^T(\mathbf{M}\mathbf{M}^T)^{-1}\mathbf{M}\mathbf{P}^T]^{-1}	[2\alpha\mathbf{1}_K+\mathbf{P}\mathbf{V}^T-\mathbf{P}\mathbf{M}^T(\mathbf{M}\mathbf{M}^T)^{-1}(2\alpha\mathbf{1}_n+\mathbf{M}\mathbf{V}^T)] \\ 
	\end{eqnarray}}
Actually the optimal $\Phi_1$ and $\Phi_2$ can be obtained by conjugate gradient method \cite{golub2013matrix}, the final solution of $\tilde{\mathbf{\Lambda}}$ can be obtained by Eq. \eqref{lambdasolution} with Eq.
	\eqref{phi}, the solution of $\mathbf{\Lambda}$ can be obtained by reshaping vector $\tilde{\mathbf{\Lambda}}$ back to matrix. 
	
\begin{figure}[h]
	%	\vspace{-3cm}
	\centering
	\includegraphics[width=15cm,height=5cm]{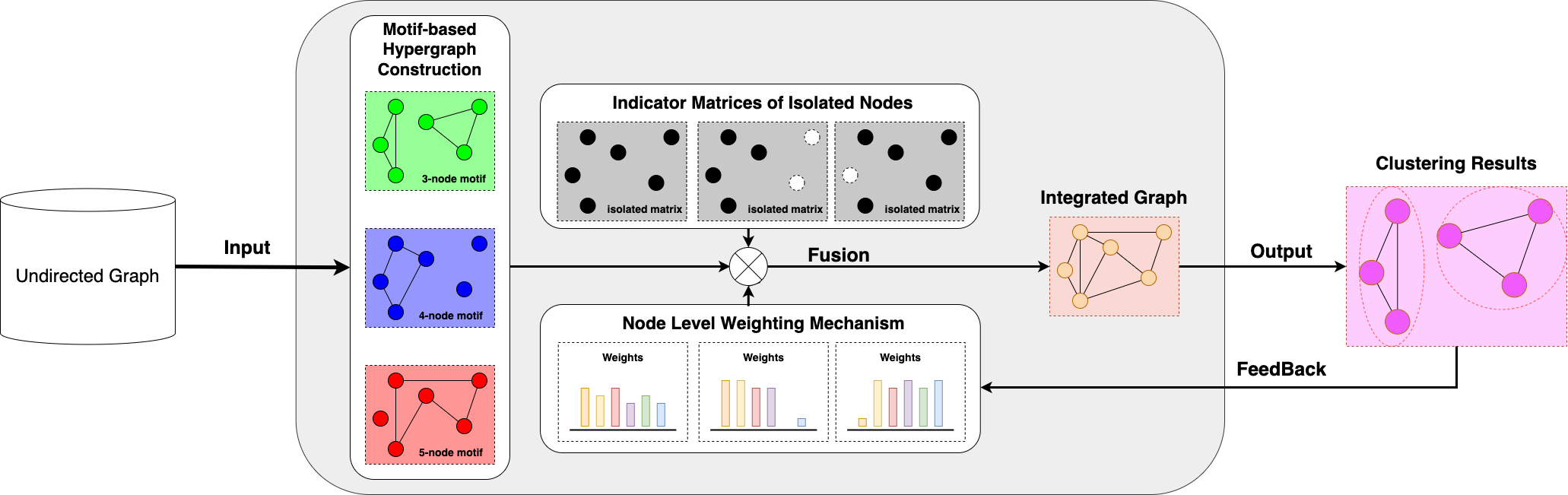}
	\caption{The flowchart of MOGC}
	\label{fig:flowchat}
\end{figure}

	Based on the above analysis, the detail process for solution Eq. \eqref{opt} is shown in the Algorithm \ref{algorithm}. Besides, the flowchart of our proposed methods is shown in Fig. \eqref{fig:flowchat}. 
	\begin{algorithm}
		\caption{MOGC: Alternating method for solving Eq. \eqref{opt}}\label{algorithm}
        \raggedright
		\textbf{Input:} Given input $\mathbf{A}_{M_{P_j}^{q_j}}$, $\mathbf{W}_{M_{P_j}^{q_j}}$, the parameters $\alpha$, $K$ and the stopping criterion $\epsilon=10^{-5}$.\\
        \textbf{Output:} $\mathbf{U}$ and $\mathbf{\Lambda}$
		\begin{algorithmic}[1]
			\REPEAT
			\STATE Compute $\mathbf{U}^k$ by (\ref{opt1}).
			\STATE Compute $\tilde{\mathbf{\Lambda}}$ by Eq. \eqref{lambdasolution}, and reshape $\tilde{\mathbf{\Lambda}}$ into $n\times n$ matrix $\mathbf{\Lambda}$.
			\UNTIL{$\max\{\|\mathbf{U}^{k+1}-\mathbf{U}^k\|_F^2, \|\mathbf{\Lambda}^{k+1}-\mathbf{\Lambda}^k\|_F^2\}\leq \epsilon$.}
		\end{algorithmic}
	\end{algorithm}
\section{Theoretical Analysis }
\subsection{Convergence Analysis}
Denote $\mathscr{F}:=\left\{({\bf \Lambda},{\bf U})|{\bf U}^{\rm T}{\bf D}_f{\bf U}={\bf I}, {\bf \Lambda}{\bf 1}_{m}={\bf 1}_n,{\bf \Lambda}_{i,j}\geq 0,
		(i,j)\in[1,n]\times [1,m]\right\}$
The minimization problem can be written as:
\begin{equation*}
	\min\limits_{({\bf \Lambda},{\bf U})\in\mathscr{F}}f({\bf \Lambda},{\bf U}),\text{  where  }
	f({\bf \Lambda},{\bf U})={\rm tr}({\bf U}^{\rm T}{\bf L}_f{\bf U})+\alpha||{\bf \Lambda}||_{F}^2.
\end{equation*}
Let $\{{\bf \Lambda}^{(k)},{\bf U}^{(k)}\}_{k=0}^{+\infty}$ be the iterative sequence generated by the alternating minimization. 

\begin{mydef}\label{altmincvgthm}
	The values of objective function $f$ along the iterative sequence $\{{\bf \Lambda}^{(k)},{\bf U}^{(k)}\}_{k=0}^{+\infty}$ is monotonically decreasing and converge to a finite value $\check{f}\geq 0$, i.e., $0 \leq \lim\limits_{k\rightarrow+\infty}f({\bf \Lambda}^{(k)},{\bf U}^{(k)})=\check{f}< +\infty$ and $\check{f}=\inf\limits_{k\geq 0}f({\bf \Lambda}^{(k)},{\bf U}^{(k)})$.
\end{mydef}

\begin{myproof}
	The alternating iterative process implies that
	\begin{equation*}
		f({\bf \Lambda}^{(k)},{\bf U}^{(k)})\geq f({\bf \Lambda}^{(k)},{\bf U}^{(k+1)})\geq f({\bf \Lambda}^{(k+1)},{\bf U}^{(k+1)}),
	\end{equation*}
	which means the sequence $\{f({\bf \Lambda}^{(k)},{\bf U}^{(k)})\}_{k=1}^{+\infty}$ is monotonically decreasing. Moreover, it is clear that $\{{\bf \Lambda}^{(k)},{\bf U}^{(k)}\}_{k=0}^{+\infty}\subset \mathscr{F}$. Therefore, $+\infty>f({\bf \Lambda}^{(k)},{\bf U}^{(k)})\\\geq 0$. Hence, $\{f({\bf \Lambda}^{(k)},{\bf U}^{(k)})\}_{k=1}^{+\infty}$ is a monotonically decreasing upper and lower bounded sequence. Hence, there exists a finite constant $\check{f}\in[0,+\infty)$ such that $\lim\limits_{k\rightarrow+\infty}f({\bf \Lambda}^{(k)},{\bf U}^{(k)})=\check{f}$. The monotonicity of $\{f({\bf \Lambda}^{(k)},{\bf U}^{(k)})\}_{k=1}^{+\infty}$  implies that  $\check{f}=\inf\limits_{k\geq 0}f({\bf \Lambda}^{(k)},{\bf U}^{(k)})$. The proof is complete.
\end{myproof}
\subsection{Computational Complexity}
We analyze the complexity of the proposed MOSC method as follows. In general, the complexity of this algorithm is governed by the computation of motif adjacency matrix and the alternating minimization algorithm in Algorithm. 1. 

{For sparse real networks with adjacency matrix $\mathbf{A}$, the symmetric matrix $\mathbf{A}$ has $|\mathbf{A}|$ non-zero elements, we assume that there is $|\mathbf{A}_i|$ non-zero elements at the $i$-th column. According to the introduction of section 3.2.1, for the motif $M_3^3$, its corresponding motif based adjacency matrix construction is $\mathbf{A}_{M_{p_j}^{q_j}}=(\mathbf{A}\circ \mathbf{A}^T)\cdot(\mathbf{A}\circ \mathbf{A}^T)\circ(\mathbf{A}\circ \mathbf{A}^T)$, where $\mathbf{A}\circ \mathbf{A}^T$ takes $|\mathbf{A}|$ flops, and the $\cdot$ operation takes $\sum_{i=1}^n|\mathbf{A}_i|^2$ flops, so the total flops of $M_3^3$ motif based adjacency matrix construction is $\sum_{i=1}^n|\mathbf{A}_i|^2+3|\mathbf{A}|$ and the corresponding computational complexity is $O(\sum_{i=1}^n|\mathbf{A}_i|^2)$. $M_3^2$ based adjacency matrix is generated according to \cite{benson2016higher}, $O(\frac{nJ_i}{2})$  complexity is required, where $J_i$ is the non-zero elements at the $i$-th row of bidirection subgraph adjacency matrix $\mathbf{\hat{B}}$ corresponding to $\mathbf{A}$.
	
For $p$-node motif with $p=4$ or $p=5$,  according to section 3.2.2, a sampling method with mFinder software is firstly applied to count all motif instances, which has total computation complexity of ${O}(sp^{p+1})$, where $s$ is the number of samples \cite{kashtan2004efficient}. Then, Algorithm 1 in \cite{zhao2019ranking} is executed to generate motif-based adjacent matrix, it has complexity ${O}(s_{p} p^2)$, where $s_{p}$ denotes the number of generated motif instances of $p$-node motif. Therefore, the total complexity for calculating 4-node or 5-node based adjacent matrix is ${O}(s p^{p+1}+s_p p^2)$.

% Theoretically, we can compute motif adjacency matrix in $\mathcal{O}(n^p)$ time for a motif with $p$ nodes and $n$ graph nodes. However, most real-world networks are sparse, we can instead focus on the computational complexity in terms of the number of edges denoted by $E$ in the network. For sparse real networks, to calculate n-node motif based adjacent matrix, many efficient practical methods have been proposed in the literature \cite{jha2015path}\cite{kashtan2004efficient}. Some methods show good performance in computation time and accuracy, for example, sampling method \cite{kashtan2004efficient} is adopted in this paper. When calculating $p$-node motif-based adjacent matrix, we firstly run the sampling method with mFinder software, which has total computation complexity of $\mathcal{O}(s\times p^{p+1})$, where $s$ is the number of samples \cite{kashtan2004efficient}. {\color{blue}Assuming the number of generated motif instances of motif $M_p^q$ is $s_{M_p^q}$}, then we execute Algorithm 1 in \cite{zhao2019ranking} to generate motif-based adjacent matrix, 
% it has complexity $\mathcal{O}(s_{M_p^q}\times n^2)$. Therefore, the total complexity for calculating adjacent matrix is $\mathcal{O}(s\times n^{n+1})+\mathcal{O}(s_{M_p^q}\times n^2)$.
Next, we discuss the operation cost of the alternating minimization iterative process when the matrices ${\bf A}_{M_{p_j}^{q_j}},~{\bf W}_{M_{p_j}^{q_j}}$ are given. For the process of solving $\mathbf{\Lambda}$ with fixed ${\bf U}$, the $\mathbf{\Lambda}$ is calculated by Eq. \eqref{lambdasolution}, which requires the value of $\Phi_1$, $\Phi_2$, $\mathbf{P}$ and $\mathbf{V}$.  By definition of $\Phi_1$ in Eq. \eqref{phi}, its evaluation requires $(2K^2+5K+1)mn+(4K^2+2K+11)n+K^3+K^2+K$ flops using conjugate gradient solver. By definition of $\Phi_2$ in Eq. \eqref{phi}, its evaluation requires $(2K^2+3K+2)mn+(2K^2-K+1)n+K^3/3+5K^2/2+K/6$ flops. Besides, both of $\mathbf{P}$ and $\mathbf{V}$ depends on $\tilde{\bf A}_i$, $\hat{\bf A}_i$ and $\bar{\bf A}_i$ for $i=1,2,...,K$. By definition of $\tilde{\bf A}_i$, its evaluation requires $(2n-1)m+2\tilde{s}$ flops for each $i$. Here, $\tilde{s}$ represents the total number of non-zero entries in  $\mathbf{A}_{{M}_{p_1}^{q_1}},\ldots, \mathbf{A}_{{M}_{p_m}^{q_m}}$. By the definition of $\hat{\bf A}_i$, its evaluation requires $2\tilde{s}+n$ flops for each $i$. By definition of $\bar{\bf A}_i$, its evaluation requires $2\tilde{s}$ flops for each $i$. By definition of ${\bf P}$, its evaluation requires $Kmn$ flops with the knowledge of $\hat{\bf A}_i$ and $\bar{\bf A}_i$. By definition of ${\bf V}$, its evaluation requires $2Kmn$ flops with the knowledge of $\hat{\bf A}_i$, $\bar{\bf A}_i$ and $\tilde{\bf A}_i$. Eq. \eqref{lambdasolution} indicates that the evaluation of ${\bf  \Lambda}$ requires $(2K+2)mn$ flops with the knowledge of $\Phi_1$, $\Phi_2$ and ${\bf V}$.
	Therefore,  the updating ${\bf \bf \Lambda}$ for fixed ${\bf U}$ requires $(4K^2+15K+5)mn+(6K^2+3K+12)n+(6\tilde{s}-m)K+4K^3/3+7K^2/2+7K/6$ flops in total, and it is an $O(K^2mn)$ sub-problem.

% $\Phi_1$ and $\Phi_2$, ${\bf \Lambda}$ is updated by \eqref{lambdasolution}. \eqref{lambdasolution} requires evaluations of ${\bf B}$, ${\bf V}$, ${\bf \Phi}_1$, ${\bf \Phi}_2$. These evaluations further require evaluations of $\tilde{\bf A}_i$, $\hat{\bf A}_i$ and $\bar{\bf A}_i$ for $i=1,2,...,K$. By the definitions of $\tilde{\bf A}_i$, $\hat{\bf A}_i$ and $\bar{\bf A}_i$, evaluations of $\tilde{\bf A}_i$, $\hat{\bf A}_i$ and $\bar{\bf A}_i$ requires $\mathcal{O}(MN)$  operations for each $i$. By definitions of ${\bf B}$, ${\bf V}$, we know that the evaluations of  ${\bf B}$ and  ${\bf V}$ require $\mathcal{O}(KMN)$ operations. By \eqref{phi}, the evaluations of $\Phi_1$ and  $\Phi_2$ requires $\mathcal{O}(K^2MN)$ using conjugate gradient solver. Therefore, the updating ${\bf \bf \Lambda}$ for fixed ${\bf U}$ requires $\mathcal{O}(K^2MN)$ operations.
Now, let's focus on the process of updating ${\bf U}$ with fixed ${\bf \Lambda}$. It is an eigen-problem for find the $K$ eigenvectors corresponding to the $K$ smallest eigenvalues of the symmetric matrix ${\bf D}^{-\frac{1}{2}}{\bf L}{\bf D}^{-\frac{1}{2}}$, which requires $\mathcal{O}(K^2n)$ operations by the truncated SVD algorithm.

 To conclude, each alternating minimization iteration requires $\mathcal{O}(K^2mn)$ operations in total.}
	\section{Experiments}
	\subsection{Testing Datasets}
	In this section, we conduct experiments on seven real datasets.	\textbf{football}\footnote[1]{http://www-personal.umich.edu/~mejn/netdata/} contains 115 nodes connected with 616 edges which are partitioned into 12 groups.
	\textbf{polbooks}\footnotemark[1] is a network of 105 books  with 441 edges, edges represent frequent copurchasing of books by the same buyers. Books belongs to 3 clusters.	
	\textbf{polblogs}\footnotemark[1] is a network with 1490 weblogs (nodes) partitioned into 2 clusters and 19090 edges.
	\textbf{Cora}\footnote[2]{https://linqs.org/datasets/} consists of 2708 scientific publications classified into one of seven classes, nodes are connected with 5429 links. 
	\textbf{email-Eu-core}\footnote[3]{https://snap.stanford.edu/data/email-Eu-core.html} describes the email sent between members (nodes) within a large European research institution. The network contains 1005 nodes with 25571 edges are classified into one of 42 classes.
	\textbf{CiteSeer}\footnotemark[2] consists of 3312 scientific publications classified into one of six classes. It has 4732 links.
	\textbf{Pubmed}\footnotemark[2] consists of 19717 scientific publications with 44338 links classified into one of three classes. All datasets are undirected graph and self-loops are removed, the summary of them is shown in Table. \ref{dataset}.
	\begin{table}[ht]
%		\vspace{-0.3cm}
		\caption{Summary of real datasets}\label{dataset}
		\centering
			\vspace{-0.2cm}
		\begin{tabular}{c|ccccccc}
			\hline
			Dataset& football&polbooks&polblogs&Cora&email-Eu-core&Citeseer&Pubmed\\
			\hline
			Nodes&115&105&1490&2708&1005&3312&19717\\
			\hline
			Edges&616&441&19090&5429&25571&4732&44324\\
			\hline
		{Class}&12&3&2&7&42&6&3\\
			\hline
		\end{tabular}
	\end{table}
	\subsection{Comparison Methods}
	We mainly compare our method with several lower-order clustering methods and higher-order clustering methods. {The lower-order clustering methods:} 
%	only focus on nodes and edges, 
	\textbf{Spectral Clustering (SC) }\cite{ng2001spectral} is performed on edge-based adjacency matrix. \textbf{Normalized Cut (Ncut)} \cite{shi2000normalized} partitions graphs based on graph normalized cut. \textbf{Nonnegative Matrix Factorization} \cite{wang2011community} is performed on the graph adjacency matrix to obtained the group partition. \textbf{Affinity Propogation (AP) } \cite{frey2007clustering} seeks cluster of nodes by similarities between pairs of nodes. \textbf{Node2vec+Kmeans (N2VKM)} \cite{grover2016node2vec} learns low-dimensional representations for vertices by biased random walk and skip-Gram, then K-means is performed to cluster the embedded vectors of vertices into several groups. 
	 {The higher-order clustering methods} 
	include Motif\_SC \cite{benson2016higher}, EdMot\_SC \cite{li2019edmot},  MWLP \cite{li2020community}, MOSC \cite{GE2021mosc}, CDMA \cite{li2023multiplex} and MotifCC \cite{wu2024motif}. They can only utilize one kind of motif, while our method integrates information from multiple motifs. 
	In the experiment, higher-order clustering methods with motif $\mathbf{M}_q^p$ means motif $\mathbf{M}_q^p$ is used to generate motif hypergraph.
\textbf{Motif\_SC} \cite{benson2016higher} extends traditional spectral clustering to higher-order structure by constructing motif-based adjacency matrix. \textbf{EdMot\_SC} \cite{li2019edmot} proposes a edge enhancement approach to overcome hypergraph fragmentation issues by adding edges to generate hypergraph,
%, the generated motif hypergraph is based on one kind of motif.
 then the new constructed adjacency matrix is used as input of spectral clustering to obtain final.
% the lower-order clustering methods. In order to compare with our method, we adopt spectral clustering in the final step.
		%	 then add edges to fragmented motif hypergraph, finally apply spectral clustering on contructed adjacency matrix.
		 \textbf{MWLP} \cite{li2020community} proposes motif-aware weighted label propagation method to integrate higher-order structure and original lower-order structure by reweighted network. 
		  \textbf{MOSC} 
 \cite{GE2021mosc} performs spectral clustering on the mixed-order adjacency matrix generated based on edge and one kind of motif by graph Laplacian. 
 {\textbf{CDMA} \cite{li2023multiplex} proposes a multiplex network community detection algorithm based on edge and motif, only one layer is considered in order to make comparison with our method.}
 {\textbf{MotifCC} \cite{wu2024motif} performs deep contrastive learning on sub-higher order graph and  sub-lower order graph, which generated from motif based graph and edge based graph separately. }   
		  	\begin{figure}
		  	\vspace{-3cm}
		  	\centering
		  	\includegraphics[width=12cm,height=3cm]{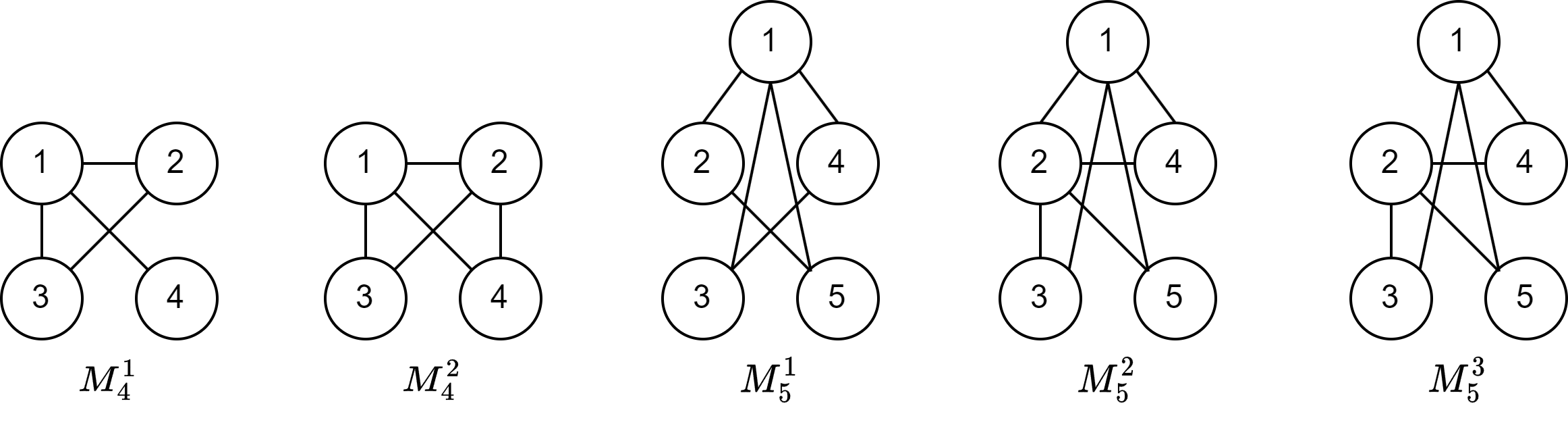}
		  	\caption{4-node motif and 5-node motif used in the experiments.}\label{motif}
		  	\vspace{-0.5cm}
		  \end{figure}
\subsection{Experiments Setting and Evaluation Measurements}
  For MOGC, $\alpha$ is tuned in $\{0.1,0.2,\ldots,3\}$ and $\{4,5,\ldots,20\}$, and then it is determined when best score is achieved.  In addition, the walk length, the number of walks per node, the embedding dimension and the window size in Node2vec are set as 80, 10, 128, and 10 separately.
%  for N2VKM, we set the number of clusters in K-means as the ground-truth communities. 
%  for the parameters in Node2vec, the walk length is 80, the number of walks per node is 10, the embedding dimension is 128, and the window size is 10.
   Moreover, the hyperparameter $p$ and in-out parameter $q$ are tuned such that best performance is achieved.  
   Best parameters reported in other comparison methods are used in the experiment. We set the number of clusters as the ground-truth communities for all methods. 
  
  	We evaluate the clustering results using metrices like Rand Index (RI) \cite{rand1971objective} and Normalized Mutual Information (NMI) \cite{amelio2015normalized}. For all measurements, a higher value indicates better performance. All experiments are repeated for 20 times,  the average metrics and their standard deviations are reported.
\subsection{Experimental Results}
\subsubsection{Results of 3-node Motifs}
We firstly test our method on 3-node motif, there are 2 different types of 3-node motif for undirected graph, i.e., ${M}_3^3$ and ${M}_3^2$ shown in Fig. \ref{motif}. Then, we 
construct motif ${M}_3^3$ and ${M}_3^2$ based adjacency matrix introduced in section 3. 
%Specifically, ${M}_3^3$ motif-based adjacency matrix is calculated using matrix computation as described in Section \ref{compute}. For ${M}_3^2$ motif-based adjacency matrix, we run Algorithm 1 in \cite{zhao2019ranking} to generate  after we obtain the bidirectional links of $\mathcal{G}$. 
The lower-order clustering methods including SC, Ncut, NMF, AP, N2VKM are tested on edge-based graph, the higher-order clustering methods utilizing one kind of motif and edge, including Motif\_SC, EdMot\_SC, MWLP, MOSC, CDMA, and MotifCC, are tested on edge-based graph and  motif-based hypergraph. Our method MOGC can utilize lower-order graph and multiple higher-order hypergraphs, the results of MOGC with ${M}_3^3$ and ${M}_3^2$ on seven datasets mean that edge-based graph, motif ${M}_3^3$ and motif ${M}_3^2$ based hypergraphs are used in Algorithm \ref{algorithm}. The experiment results for all methods on seven datasets are shown in Table. \ref{resultRI} and Table. \ref{resultnmi}, it shows that MOGC outperforms other methods with different motifs, which demonstrates the effectiveness of our method.
\begin{table}
	\vspace{-4cm}
	\caption{Rand Index Results on seven real datasets (mean$\pm$std).}\label{resultRI}
	\centering
	\scalebox{0.63}{\begin{tabular}{c|cccccccc}
			\hline
			&dataset&football&polbooks&polblogs&Cora&email-Eu-core&Citeseer&Pubmed\\
			\hline
			&SC&{\small {0.8967\scriptsize{$\pm$0.0000}}}&	{\small {0.6445\scriptsize{$\pm$0.0521}} }&	{\small {0.0005 \scriptsize{$\pm$0.0000}} }&	{\small {0.2385  \scriptsize{$\pm$0.0000}}  } &	{\small {0.4145\scriptsize{$\pm$0.0188} }  }&	{\small {0.3431\scriptsize{$\pm$0.0000}} }&	{\small {0.1123 \scriptsize{$\pm$0.0000}} }\\
			\cline{2-9}					
			&Ncut&{\small  {0.0086 \scriptsize{$\pm$0.0052}} }&{\small {0.3524 \scriptsize{$\pm$0.0782}} }&{\small  {0.3497 \scriptsize{$\pm$0.0441 }} }&{\small {0.0027\scriptsize{$\pm$0.0016}} }&{\small  {0.0260 \scriptsize{$\pm$0.0050}}  }&{\small  {0.0638 \scriptsize{$\pm$0.0145 }}} &{\small  {0.0004\scriptsize{$\pm$0.0008}} }\\
			\cline{2-9}
			edge&NMF&{\small {0.8810\scriptsize{$\pm$0.0254}} }& {\small {0.5679\scriptsize{$\pm$0.0523}}} & {\small  {0.8013 \scriptsize{$\pm$0.0000}} }&{\small {0.2745\scriptsize{$\pm$0.0209}} }&{\small {0.4574\scriptsize{$\pm$0.0200}}  }&{\small {0.1584\scriptsize{$\pm$0.0108} }}&{\small {0.0976 \scriptsize{$\pm$0.0039}}}\\
			\cline{2-9}
			&AP&{\small  {0.3426\scriptsize{$\pm$0.0460} }}&{\small  {0.1139 \scriptsize{$\pm$0.0171}}} &{\small {0.0501 \scriptsize{$\pm$0.0021 }} }&{\small {0.0160 \scriptsize{$\pm$0.0011}}}&{\small {0.0739\scriptsize{$\pm$0.0067} } }&{\small  {0.0117\scriptsize{$\pm$0.0015}}}&{\small {0.0011 \scriptsize{$\pm$0.0002}}}\\
			\cline{2-9}
			&N2VKM&{\small {0.9836 \scriptsize{$\pm$0.0000}} }&{\small {0.8485\scriptsize{$\pm$0.0000}}}&{\small {0.8186 \scriptsize{$\pm$0.0000}}}&{\small {0.8241 \scriptsize{$\pm$0.0000}}}&{\small {0.9553 \scriptsize{$\pm$ 0.0000}}}&{\small {0.7182 \scriptsize{$\pm$ 0.0005}}}&{\small {0.6905 \scriptsize{$\pm$0.0000}}}\\
			\hline
			&Motif\_SC&{\small {0.8967 \scriptsize{$\pm$0.0273}} } &{\small  {0.6572 \scriptsize{$\pm$0.0713}} } &{\small {0.0036 \scriptsize{$\pm$0.0000 }} } &{\small {0.0788 \scriptsize{$\pm$0.0000}}} &{\small {0.3243 \scriptsize{$\pm$0.0104} } }&{\small {0.0369 \scriptsize{$\pm$0.0000}} }&{\small {0.3511 \scriptsize{$\pm$0.0078}} }\\
			\cline{2-9}
			&EdMot\_SC&{\small {0.8967 \scriptsize{$\pm$0.0000} } }&{\small {0.6622 \scriptsize{$\pm$0.0236} } }&{\small {0.7956 \scriptsize{$\pm$0.0000} } } &{\small {0.3434 \scriptsize{$\pm$0.0050}}} &{\small {0.3667 \scriptsize{$\pm$0.0119}}  }&{\small {0.1545 \scriptsize{$\pm$0.0342}} }&{\small {0.2623 \scriptsize{$\pm$0.0000}} }\\
			\cline{2-9}
			${M}_3^3$&MWLP&{\small {0.9420 \scriptsize{$\pm$0.0010}} } &{\small {0.6689\scriptsize{$\pm$0.0138}} }&{\small {0.5277 \scriptsize{$\pm$0.0018}} } & {\small {0.7761 \scriptsize{$\pm$0.0024}} }&{\small {0.1902 \scriptsize{$\pm$0.0000}}  }&{\small {0.7813 \scriptsize{$\pm$0.0004}} }&{\small {0.6405 \scriptsize{$\pm$0.0011}}} \\
			\cline{2-9}
			&MOSC&{\small {0.7733 \scriptsize{$\pm$0.0596}}  }&{\small {0.6613 \scriptsize{$\pm$0.0073} }} &{\small {0.0004 \scriptsize{$\pm$0.0000}  }}  &{\small {0.0053 \scriptsize{$\pm$0.0000}}} &{\small {0.4718  \scriptsize{$\pm$0.0211}} } &{\small {  0.3225 \scriptsize{$\pm$0.0224}}} &{\small {0.1880 \scriptsize{$\pm$0.0762}} }\\
			\cline{2-9}			&CDMA&	{\small { 0.9597\scriptsize{$\pm$0.0000}} }&{\small { 0.4844\scriptsize{$\pm$0.0000}}} &{\small { 0.5195\scriptsize{$\pm$0.0000} } } &{\small {0.1999\scriptsize{$\pm$0.0000}}} &{\small {0.7859\scriptsize{$\pm$0.0000}}  }&{\small { 0.2169\scriptsize{$\pm$0.0000} }} &{\small {0.3580\scriptsize{$\pm$0.0000}}}\\	\cline{2-9}
			&MotifCC&{\small {0.9706 \scriptsize{$\pm$0.0063}} } &{\small  {0.6774 \scriptsize{$\pm$0.0563}} } &{\small {0.7131 \scriptsize{$\pm$0.0562 }} } &{\small {0.7334 \scriptsize{$\pm$0.0015}}} &{\small {0.9409 \scriptsize{$\pm$0.0035} } }&{\small {0.7147 \scriptsize{$\pm$0.0006}} }&{\small {0.5479 \scriptsize{$\pm$0.0003}} }\\
			\cline{2-9}
			&MOGC&{\small {0.9830 \scriptsize{$\pm$0.0047}} }&{\small {0.8464 \scriptsize{$\pm$0.0035}}} &{\small {0.8963 \scriptsize{$\pm$0.0000} } } &{\small {0.7375\scriptsize{$\pm$0.0000}}} &{\small {0.9564\scriptsize{$\pm$0.0102}}  }&{\small {0.6965 \scriptsize{$\pm$0.0000} }} &{\small {\textbf{0.6510} \scriptsize{$\pm$0.0081}}}\\ 
			\hline
			&Motif\_SC&{\small {0.8609 \scriptsize{$\pm$0.0751} }}&{\small {0.6636 \scriptsize{$\pm$0.0451} }}&{\small {0.8190 \scriptsize{$\pm$0.0000} }}&{\small {0.2726 \scriptsize{$\pm$0.0000}}}&{\small {0.4431 \scriptsize{$\pm$0.0125}} }&{\small {0.3461 \scriptsize{$\pm$0.0000}}}&{\small {0.2597 \scriptsize{$\pm$0.0000} } } \\
			\cline{2-9}
			&EdMot\_SC&{\small  {0.8590\scriptsize{$\pm$0.0000} }}&{\small {0.6667 \scriptsize{$\pm$0.0080} }}&{\small  {0.8072 \scriptsize{$\pm$0.0000}} }&{\small {0.2726 \scriptsize{$\pm$0.0000}}}&{\small {0.4414 \scriptsize{$\pm$0.0067} }}&{\small {0.3461 \scriptsize{$\pm$0.0000}}}&{\small {0.2597 \scriptsize{$\pm$0.0000}} }\\
			\cline{2-9}
			${M}_3^2$&MWLP&{\small {0.8900\scriptsize{$\pm$0.0015} }}&{\small {0.6559 \scriptsize{$\pm$0.0007} } }&{\small {0.5847 \scriptsize{$\pm$0.0000} }}&{\small {0.7793 \scriptsize{$\pm$0.0006}}}&{\small {0.1485 \scriptsize{$\pm$0.0000}}}&{\small {0.7468 \scriptsize{$\pm$0.0000}}}&{\small {0.6405 \scriptsize{$\pm$0.0013}}}\\
			\cline{2-9}
			&MOSC&{\small {0.7376 \scriptsize{$\pm$0.0724} }}&{\small {0.6326\scriptsize{$\pm$0.0070} } } &{\small {0.8190 \scriptsize{$\pm$0.0000} } }&{\small {0.0055 \scriptsize{$\pm$0.0000}} }&{\small {0.4399 \scriptsize{$\pm$0.0233}}}&{\small {0.3212\scriptsize{$\pm$0.0214}}}&{\small{0.2376 \scriptsize{$\pm$0.0495}} }\\
			\cline{2-9}
			&CDMA&	{\small {0.9408\scriptsize{$\pm$0.0000}} }&{\small{0.7170\scriptsize{$\pm$0.0000}}} &{\small{0.4996\scriptsize{$\pm$0.0000}}} &{\small{0.4522\scriptsize{$\pm$0.0000}}} &{\small{0.8306\scriptsize{$\pm$0.0000}}}&{\small{0.3903\scriptsize{$\pm$0.0000}}} &{\small{0.3995\scriptsize{$\pm$0.0000}}}\\ \cline{2-9}		&MotifCC&{\small {0.9617 \scriptsize{$\pm$0.0090}} } &{\small  {0.7807 \scriptsize{$\pm$0.0390}} } &{\small {0.6838 \scriptsize{$\pm$0.0435 }} } &{\small {0.7351 \scriptsize{$\pm$0.0016}}} &{\small {0.9355 \scriptsize{$\pm$0.0032} } }&{\small {0.7155 \scriptsize{$\pm$0.0006}} }&{\small {0.5454 \scriptsize{$\pm$0.0002}} }\\
			\cline{2-9}	
			&MOGC&{\small {\textbf{0.9858} \scriptsize{$\pm$0.0108}} }&{\small {\textbf{0.8555} \scriptsize{$\pm$0.0000}}}&{\small {\textbf{0.9110} \scriptsize{$\pm$0.0000}} }&{\small {0.7888  \scriptsize{$\pm$0.0000}} }&{\small {0.9608 \scriptsize{$\pm$0.0009}} }&{\small {{0.7883} \scriptsize{$\pm$0.0000 }} }&{\small {0.5405 \scriptsize{$\pm$0.0015}} }\\
			\hline
			
			&Motif\_SC&{\small {0.8242  \scriptsize{$\pm$0.0000}} }&{\small {0.6570  \scriptsize{$\pm$0.0000} } }&{\small {0.0509   \scriptsize{$\pm$0.0000} } }&{\small {0.1777 \scriptsize{$\pm$0.0000} } }&{\small {0.3858 \scriptsize{$\pm$0.0000} } }&{\small {0.0408 \scriptsize{$\pm$0.0000} } }&{\small{0.0007  \scriptsize{$\pm$0.0000} } }\\
			\cline{2-9}
			&EdMot\_SC&{\small{ 0.8242 \scriptsize{$\pm$0.0000}} }&{\small {0.6886  \scriptsize{$\pm$0.0000}} }&{\small {0.1628  \scriptsize{$\pm$0.0818} } }&{\small {{0.3453} \scriptsize{$\pm$0.0082} }}&{\small  {{0.3933} \scriptsize{$\pm$0.0036} } }&{\small{ 0.1211 \scriptsize{$\pm$0.0425}}  }&{\small { 0.0626 \scriptsize{$\pm$0.0003}}}\\
			\cline{2-9}
			${M}_{4}^*$&MWLP&{\small {{0.9396} \scriptsize{$\pm$0.0009} } }&{\small {{0.8209} \scriptsize{$\pm$0.0141} } }&{\small {{0.5083} \scriptsize{$\pm$0.0010} }  } & {\small {{0.8024} \scriptsize{$\pm$0.0005} }  }&{\small {{ 0.9164} \scriptsize{$\pm$0.0001} }  }&{\small  {\textbf{ 0.8222 } \scriptsize{$\pm$0.0000} }  }& {\small {{ 0.6400 } \scriptsize{$\pm$0.0000} } }\\
			\cline{2-9}
			&MOSC& {\small {{0.7526} \scriptsize{$\pm$0.0472} }}&{\small {{0.6484} \scriptsize{$\pm$0.0086} } }&{\small {{0.7371} \scriptsize{$\pm$0.2458} } }  &{\small {{0.0544} \scriptsize{$\pm$0.0028} } } &{\small {{ 0.3838} \scriptsize{$\pm$0.0149} }  }&{\small  {{ 0.0180 } \scriptsize{$\pm$0.0028} } } &{\small \makecell{{ 0.1017 } \scriptsize{$\pm$0.0035} } }\\
			\cline{2-9}
			&CDMA&	{\small {0.9553\scriptsize{$\pm$0.0000}} }&{\small{0.4727\scriptsize{$\pm$0.0000}}} &{\small{0.5015\scriptsize{$\pm$0.0000}}} &{\small{0.1944\scriptsize{$\pm$0.0000}}} &{\small{0.8114\scriptsize{$\pm$0.0000}}}&{\small{0.2071\scriptsize{$\pm$0.0000}}} &{\small{0.3607\scriptsize{$\pm$0.0000}}}\\ \cline{2-9}
			&MotifCC&{\small {0.9712 \scriptsize{$\pm$0.0098}} } &{\small  {0.7224 \scriptsize{$\pm$0.0598}} } &{\small {0.6394 \scriptsize{$\pm$0.0425 }} } &{\small {0.7333 \scriptsize{$\pm$0.0014}}} &{\small {0.9433 \scriptsize{$\pm$0.0032} } }&{\small {0.7147 \scriptsize{$\pm$0.0003}} }&{\small {0.5478 \scriptsize{$\pm$0.0002}} }\\
			\cline{2-9}
			&MOGC&{\small {0.9759  \scriptsize{$\pm$0.0000}} }&{\small {0.8463  \scriptsize{$\pm$0.0081}} }&{\small {0.9036 \scriptsize{$\pm$0.0000}}}&{\small {\textbf{0.8196}  \scriptsize{$\pm$0.0000}} }&{\small {0.9570 \scriptsize{$\pm$0.0008}}}&{\small {0.7754 \scriptsize{$\pm$0.0000}}}&{\small {0.5855  \scriptsize{$\pm$0.0000}}}\\
			\cline{2-9}
			\hline
			&Motif\_SC& {\small \makecell{0.8255  \scriptsize{$\pm$0.0000}}}&{\small {0.5264 \scriptsize{$\pm$0.0000}}} &  {\small {0.5065 \scriptsize{$\pm$0.0000}} }&{\small {0.0037 \scriptsize{$\pm$0.0000}}}& {\small {0.1490 \scriptsize{$\pm$0.0000}} }&{\small  {0.0082 \scriptsize{$\pm$0.0000}}}&{\small{0.0019  \scriptsize{$\pm$0.0000}} }\\
			\cline{2-9}
			&EdMot\_SC&{\small {0.8419 \scriptsize{$\pm$0.0336}} }&{\small {0.3671  \scriptsize{$\pm$0.1539}} }&{\small {0.8015  \scriptsize{$\pm$0.0006}}} &{\small {0.2215  \scriptsize{$\pm$0.0327}} }&{\small { 0.2858  \scriptsize{$\pm$0.0070}} }&{\small {   0.0223  \scriptsize{$\pm$0.0069}} }&{\small {    0.0906   \scriptsize{$\pm$0.0147}}} \\
			\cline{2-9}
			${M}_{5}^*$&MWLP& {\small {0.9068 \scriptsize{$\pm$ 0.0013}} }&{\small {0.7283  \scriptsize{$\pm$0.0137 }} } & {\small { 0.5151   \scriptsize{$\pm$ 0.0001}}   }&{\small  { 0.7796   \scriptsize{$\pm$ 0.0059}} }&{\small {  0.3941   \scriptsize{$\pm$0.0592 }} }&{\small  {  0.8218   \scriptsize{$\pm$0.0002 }} }&{\small { 0.6391   \scriptsize{$\pm$ 0.0061}} }\\
			\cline{2-9}
			&MOSC& {\small {0.7619 \scriptsize{$\pm$0.0784  }} }&{\small {    0.6245   \scriptsize{$\pm$   0.0057  }}  }&{\small {  0.8190    \scriptsize{$\pm$  0.0000 }}   }&{\small  { 0.0017    \scriptsize{$\pm$     0.0059 }} }&{\small {     0.3585      \scriptsize{$\pm$  0.0248  }} }&{\small  {  0.0225    \scriptsize{$\pm$   0.0004  }} }&{\small {    0.1124   \scriptsize{$\pm$ 0.0002 }} }\\
			\cline{2-9}
			&CDMA&	{\small {0.9613\scriptsize{$\pm$0.0000}} }&{\small{0.4533\scriptsize{$\pm$0.0000}}} &{\small{0.5114\scriptsize{$\pm$0.0000}}} &{\small{0.2365\scriptsize{$\pm$0.0000}}} &{\small{0.7871\scriptsize{$\pm$0.0000}}}&{\small{0.2037\scriptsize{$\pm$0.0000}}} &{\small{0.3584\scriptsize{$\pm$0.0000}}}\\ \cline{2-9}		
			&MotifCC&{\small {0.9635 \scriptsize{$\pm$0.0120}} } &{\small  {0.7585 \scriptsize{$\pm$0.0459}} } &{\small {0.7052 \scriptsize{$\pm$0.0996 }} } &{\small {0.7312 \scriptsize{$\pm$0.0021}}} &{\small {0.9400 \scriptsize{$\pm$0.0038} } }&{\small {0.7145 \scriptsize{$\pm$0.0003}} }&{\small {0.5477 \scriptsize{$\pm$0.0001}} }\\
			\cline{2-9}
			&MOGC&{\small {0.9806 \scriptsize{$\pm$0.0000} } }&{\small {0.8507 \scriptsize{$\pm$0.0000}} }&{\small {0.9021 \scriptsize{$\pm$0.0000}} }&{\small {0.7204 \scriptsize{$\pm$0.0000}}}&{\small {0.9114  \scriptsize{$\pm$0.0101}} }&{\small {0.7710 \scriptsize{$\pm$0.0000}}}&{\small {0.5377 \scriptsize{$\pm$0.0000}}}\\
			\hline
			&MOGC\_${M}_3^3$\_${M}_3^2$&{\small {{0.9847} \scriptsize{$\pm$0.0060} }}&{\small {0.8548 \scriptsize{$\pm$0.0068} } }&{\small {\textbf{0.9110} \scriptsize{$\pm$0.0000} }}&{\small {{0.7959} \scriptsize{$\pm$0.0000} }}&{\small {\textbf{0.9615} \scriptsize{$\pm$0.0009}} }&{\small {0.7865 \scriptsize{$\pm$0.0216} }}&{\small {0.5412 \scriptsize{$\pm$0.0000}}}\\
			\cline{2-9}
			{Multiple}&MOGC\_${M}_3^3$\_${M}_{4}^*$&{\small {0.9771 \scriptsize{$\pm$0.0000}} }&{\small {0.8530 \scriptsize{$\pm$0.0037}} }&{\small {0.9095 \scriptsize{$\pm$0.0000}} }&{\small {0.8048 \scriptsize{$\pm$0.0000}} }&{\small {0.9568 \scriptsize{$\pm$0.0009}} }&{\small {0.7866 \scriptsize{$\pm$0.0000}} }&{\small {0.5750 \scriptsize{$\pm$0.0000} } }\\
			\cline{2-9}
			motif&MOGC\_${M}_3^3$\_${M}_{4}^*$\_${M}_{5}^*$&{\small {0.9844 \scriptsize{$\pm$0.0000} } }&{\small {0.8519 \scriptsize{$\pm$0.0062}} }&{\small {0.9095 \scriptsize{$\pm$0.0000}}}&{\small {0.8066 \scriptsize{$\pm$0.0000}} }&{\small {0.9568 \scriptsize{$\pm$0.0009}} }&{\small {0.7869 \scriptsize{$\pm$0.0000}} }&{\small {0.5656 \scriptsize{$\pm$0.0085}} }\\
			\hline
	\end{tabular}}
\end{table}

We further give a detailed analysis about the results. Firstly, Compared with traditional edge-based graph partition methods, the corresponding higher-order methods generally have higher accuracy than most of them, which illustrates that the higher-order structures play an important role in community discovery.
Secondly, when motif ${M}_3^3$ is adopted, MOGC has better performances than other higher-order graph clustering methods on almost all datasets. For Cora, Citeseer and Pubmed datasets with very sparse connectivity pattern, the accuracy of higher-order method Motif\_SC decreases greatly compared with edge-based method (SC), since Motif\_SC suffers from serious fragmentation issue. However, in terms of Rand Index, MOGC achieves about 70\%, 65\%, 65\% improvements over Motif\_SC. This shows the effectiveness of MOGC in addressing the hypergraph fragmentation issue. {Moreover, the accuracy of CDMA decreases significantly for Cora and Citeseer datasets with $M_3^3$, due to the ignorance of fragmentation issue in CDMA.} Compared with state-of-the-art defragmented higher-order graph clustering methods (EdMot\_SC and MWLP), we also have much higher accuracy than others, which validates adaptive weights learning mechanism is more helpful than edge enchancement approaches in addressing hypergraph fragmentation issue. 
 Compared with MOSC which combines edge and triangle adjacency matrices with mixing parameter, MOGC performs better on all datasets, it proves that more accurate results would be achieved by assigning weighting scaler to each motif for each node. {Our method show better performances than MotifCC on almost all datasets, since MotifCC adopts two steps for community detection, label of all nonisolated nodes are obtained firstly by applying contrastive learning on higher-order and lower-order graph, then labels of isolated nodes are achieved by the label propagation on the original edge-based graph. While in our method, the label partitions of isolated and nonisolated nodes are realized within a optimization function, such that they can promote each other. }Finally, when motif ${M}_3^2$ is adopted, the accuracy of MOGC is further improved than that using motif ${M}_3^3$. Specially, our method has higher value of all measurements than other methods on seven datasets. This might be the fact that the density of
hypergraph using motif ${M}_3^2$ is larger than those using motif ${M}_3^3$, only few ${M}_3^2$ based hypergraphs suffer from fragmentation issue.
%In summary, the results using 3-node motif show that the integration of higher-order structure and lower-order structure at the level of node can significantly improve the performance of graph clustering. Besides, it can also tackle hypergraph fragmentation issue successfully.

\begin{table}
	\vspace{-4cm}
	\caption{NMI Results on seven real datasets (mean$\pm$std).}\label{resultnmi}
	\centering
	\scalebox{0.63}{\begin{tabular}{c|cccccccc}
			\hline
			&dataset&football&polbooks&polblogs&Cora&email-Eu-core&Citeseer&Pubmed\\
			\hline
			&SC&{\small {{0.9242}\scriptsize{$\pm$0.0000}} }&{\small {0.5422 \scriptsize{$\pm$ 0.0261}} }&{\small {0.0029  \scriptsize{$\pm$0.0000}} }&{\small  {0.3928 \scriptsize{$\pm$0.0000}} }&{\small  {  {0.6983}     \scriptsize{$\pm$0.0065 }} }&{\small  { 0.3701  \scriptsize{$\pm$0.0000}} }&{\small {0.1568 \scriptsize{$\pm$0.0000}} }\\
			\cline{2-9}
			&Ncut&{\small { 0.2435  \scriptsize{$\pm$0.0135}}}&{\small {0.3544 \scriptsize{$\pm$0.0388}} }&{\small {0.4166 \scriptsize{$\pm$0.0261}} }&{\small {0.0114 \scriptsize{$\pm$0.0037}} }&{\small  { 0.2389 \scriptsize{$\pm$0.0055}} }&{\small { 0.0718 \scriptsize{$\pm$ 0.0089}}}&{\small {0.0002  \scriptsize{$\pm$0.0001}}  }\\
			\cline{2-9}
			edge&NMF&{\small {0.9199 \scriptsize{$\pm$0.0085}} }&{\small   {0.5247 \scriptsize{$\pm$0.0127}} }&{\small {0.7147 \scriptsize{$\pm$0.0000}} }&{\small {0.3740 \scriptsize{$\pm$0.0173}} }&{\small { 0.7005	\scriptsize{$\pm$0.0053}} }&{\small  {0.2255 \scriptsize{$\pm$  0.0100}} }&{\small {0.1522 \scriptsize{$\pm$0.0062}} }\\
			\cline{2-9}
			&AP& {\small {0.6534 \scriptsize{$\pm$ 0.0295 }} }&{\small   {0.3546 \scriptsize{$\pm$ 0.0182}} }&{\small {0.2106 \scriptsize{$\pm$0.0029}} }&{\small {0.3694 \scriptsize{$\pm$0.0022}} }& {\small { 0.5402 \scriptsize{$\pm$ 0.0081}} }&{\small {0.3263 \scriptsize{$\pm$0.0029}}}&{\small{0.1637  \scriptsize{$\pm$0.0357}} }\\
			\cline{2-9}
			&N2VKM& {\small {0.9267   \scriptsize{$\pm$0.0000}} }&{\small {0.5787 \scriptsize{$\pm$0.0000}} }& {\small {0.5501 \scriptsize{$\pm$0.0000}} }&{\small {0.4643 \scriptsize{$\pm$0.0001}} }&{\small {0.7026 \scriptsize{$\pm$ 0.0000}} }&{\small {0.2498 \scriptsize{$\pm$0.0001}} }&{\small  {0.2986 \scriptsize{$\pm$0.0000}} }\\
			\hline
			&Motif\_SC&{\small {0.9242 \scriptsize{$\pm$ 0.0075}} }&{\small {0.5531 \scriptsize{$\pm$  0.0452}} } &{\small {0.0041 \scriptsize{$\pm$0.0000}} }&{\small {0.1550 \scriptsize{$\pm$0.0000}} }&{\small   {0.6498 \scriptsize{$\pm$ 0.0066 }} }&{\small  {0.0774 \scriptsize{$\pm$0.0000}} }&{\small {0.0026 \scriptsize{$\pm$0.0000}} }\\
			\cline{2-9}
			&EdMot\_SC&{\small {{0.9242} \scriptsize{$\pm$0.0000}} }&{\small {0.5583 \scriptsize{$\pm$   0.0236}} }&{\small {0.7013 \scriptsize{$\pm$0.0000}} }&{\small {0.4302 \scriptsize{$\pm$0.0020}} }&  {\small { 0.6765 \scriptsize{$\pm$ 0.0067}} }&{\small {0.2500 \scriptsize{$\pm$0.0290}} }&{\small {0.2363 \scriptsize{$\pm$0.0000}} }\\
			\cline{2-9}
			${M}_3^3$&MWLP&{\small {0.7885 \scriptsize{$\pm$0.0031}} }&{\small {0.3641 \scriptsize{$\pm$0.0134}} }&{\small {0.1397 \scriptsize{$\pm$0.0033 }} }&{\small {0.3302 \scriptsize{$\pm$0.0015}} }&{\small  {0.1053 \scriptsize{$\pm$0.0000}} }&{\small {0.0982 \scriptsize{$\pm$ 0.0005}} }&{\small {0.1557  \scriptsize{$\pm$0.0022}} }\\
			\cline{2-9}
			&MOSC&{\small {0.8837 \scriptsize{$\pm$0.0220}} }&{\small {0.5564 \scriptsize{$\pm$0.0062}} } &{\small {0.1799 \scriptsize{$\pm$0.0000}} }&{\small {0.0774 \scriptsize{$\pm$0.0000}} }&  {\small {0.6974 \scriptsize{$\pm$0.0057}} }&{\small {0.3607 \scriptsize{$\pm$ 0.0187}} }&{\small {0.2022  \scriptsize{$\pm$0.0438}} } \\ \cline{2-9}	&CDMA&	{\small {0.8675\scriptsize{$\pm$0.0000}} }&{\small{0.2000\scriptsize{$\pm$0.0000}}} &{\small{0.1109\scriptsize{$\pm$0.0000}}} &{\small{0.0193\scriptsize{$\pm$0.0000}}} &{\small{0.4593\scriptsize{$\pm$0.0000}}}&{\small{0.0350\scriptsize{$\pm$0.0000}}} &{\small{0.0030\scriptsize{$\pm$0.0000}}}\\	
			\cline{2-9}
			&MotifCC&{\small {0.8922 \scriptsize{$\pm$0.0116}} } &{\small  {0.5853 \scriptsize{$\pm$0.1027}} } &{\small {0.7414 \scriptsize{$\pm$0.0927}} } &{\small {0.0290 \scriptsize{$\pm$0.0058}}} &{\small {0.5152 \scriptsize{$\pm$0.0378} } }&{\small {0.0059 \scriptsize{$\pm$0.0017}} }&{\small {0.0006 \scriptsize{$\pm$0.0004}} }\\
			\cline{2-9}
			&MOGC&{\small {0.9214 \scriptsize{$\pm$ 0.0111}} }&{\small {0.5689 \scriptsize{$\pm$  0.0066}} }&{\small {0.7074 \scriptsize{$\pm$0.0000}} }&{\small  {0.2427 \scriptsize{$\pm$0.0000}} }&{\small {0.6860 \scriptsize{$\pm$0.0694}} }&{\small {0.2880 \scriptsize{$\pm$0.0000}} } &{\small {\textbf{0.2693} \scriptsize{$\pm$0.0041}} }\\
			\hline
			&Motif\_SC& {\small {0.8957 \scriptsize{$\pm$0.0377}} }&{\small {0.5725 \scriptsize{$\pm$0.0268} } } &{\small {0.7306 \scriptsize{$\pm$0.0000}} }&{\small {0.4355 \scriptsize{$\pm$0.0000}} }& {\small{0.6852 \scriptsize{$\pm$0.0055}} }&{\small  {0.3680 \scriptsize{$\pm$0.0000}} }&{\small {0.2508 \scriptsize{$\pm$0.0004}} }\\
			\cline{2-9}
			&EdMot\_SC&{\small	{0.9043 \scriptsize{$\pm$0.0000}} }&{\small {0.5794 \scriptsize{$\pm$ 0.0068}} }&{\small {0.7151 \scriptsize{$\pm$0.0000}} }&{\small {0.4355 \scriptsize{$\pm$0.0000}}}&{\small   {0.6878 \scriptsize{$\pm$   0.0035}} }&{\small {0.3680  \scriptsize{$\pm$0.0000}} }&{\small {0.2508  \scriptsize{$\pm$0.0000}} }\\
			\cline{2-9}
			${M}_3^2$&MWLP&{\small {0.4051 \scriptsize{$\pm$ 0.0063}}}&{\small  {0.3525 \scriptsize{$\pm$ 0.0004}} }&{\small {0.2324 \scriptsize{$\pm$0.0000}} }&{\small {0.3233 \scriptsize{$\pm$0.0017}} }&{\small {0.0531 \scriptsize{$\pm$0.0000}} }&{\small  {0.0987 \scriptsize{$\pm$0.0000}} }&{\small {0.1564 \scriptsize{$\pm$0.0014}} }\\
			\cline{2-9}
			&MOSC&{\small {0.8379 \scriptsize{$\pm$ 0.0388}} }&{\small  {0.5578 \scriptsize{$\pm$ 0.0055}} } &{\small {0.7380 \scriptsize{$\pm$0.0000}} }&{\small {0.1383 \scriptsize{$\pm$0.0000}} }&{\small {0.6791 \scriptsize{$\pm$0.0061}} }&{\small  {0.3458 \scriptsize{$\pm$0.0105}} }&{\small {0.2358 \scriptsize{$\pm$0.0329}} }\\ \cline{2-9}	
			&CDMA&	{\small {0.7811\scriptsize{$\pm$0.0000}} }&{\small{0.3974\scriptsize{$\pm$0.0000}}} &{\small{0.0003\scriptsize{$\pm$0.0000}}} &{\small{0.2129\scriptsize{$\pm$0.0000}}} &{\small{0.1288\scriptsize{$\pm$0.0000}}}&{\small{0.1480\scriptsize{$\pm$0.0000}}} &{\small{0.0525\scriptsize{$\pm$0.0000}}}\\ \cline{2-9}		
			&MotifCC&{\small {0.8518 \scriptsize{$\pm$0.0269}} } &{\small  {0.5875 \scriptsize{$\pm$0.0646}} } &{\small {0.6995 \scriptsize{$\pm$0.0707 }} } &{\small {0.0377 \scriptsize{$\pm$0.0085}}} &{\small {0.4200 \scriptsize{$\pm$0.0153} } }&{\small {0.0084 \scriptsize{$\pm$0.0025}} }&{\small {0.0007 \scriptsize{$\pm$0.0003}} }\\
			\cline{2-9}
			&MOGC&{\small {\textbf{0.9314} \scriptsize{$\pm$0.0414}} }&{\small {\textbf{0.6058} \scriptsize{$\pm$0.0000}} }&{\small {0.7338 \scriptsize{$\pm$0.0000}} }&{\small {\textbf{0.4775} \scriptsize{$\pm$0.0000}} }&{\small {0.6880 \scriptsize{$\pm$ 0.0062}} }&{\small  { 0.3734 \scriptsize{$\pm$0.0000}} }&{\small {0.1756 \scriptsize{$\pm$0.0011}} }\\
			\hline
			
			&Motif\_SC&{\small {0.9003  \scriptsize{$\pm$0.0000}} }&{\small {0.5742 \scriptsize{$\pm$0.0000}} }&{\small{0.0466 \scriptsize{$\pm$0.0000}}}&{\small {0.2843 \scriptsize{$\pm$0.0000}} }&{\small {0.6936 \scriptsize{$\pm$0.0000}} }&{\small {0.0618 \scriptsize{$\pm$0.0000}}}&{\small{0.0070 \scriptsize{$\pm$0.0000}} }\\
			\cline{2-9}
			&EdMot\_SC&{\small {0.9043 \scriptsize{$\pm$0.0000}}}&{\small {0.5761  \scriptsize{$\pm$0.0000}} }&{\small  {0.1371 \scriptsize{$\pm$0.0691}} }&{\small {0.4021 \scriptsize{$\pm$0.0078}} }&{\small {0.6975 \scriptsize{$\pm$0.0012}} }&{\small {0.1213 \scriptsize{$\pm$0.0377}} }&{\small { 0.0609 \scriptsize{$\pm$0.0002}} }\\
			\cline{2-9}
			${M}_{4}^*$&MWLP& {\small { 0.7692 \scriptsize{$\pm$0.0031}} }&{\small  { 0.5349 \scriptsize{$\pm$0.0142}} }&{\small {  0.1046  \scriptsize{$\pm$0.0040}}}&{\small  {  0.3605  \scriptsize{$\pm$0.0011}} }&{\small {  0.4488  \scriptsize{$\pm$0.0011}} }&{\small {  0.3308  \scriptsize{$\pm$0.0006}}}&{\small{  0.1517  \scriptsize{$\pm$0.0005}} }\\
			\cline{2-9}
			&MOSC& {\small { 0.8744 \scriptsize{$\pm$0.0173}} }&{\small  {0.5688 \scriptsize{$\pm$0.0055}} }& {\small { 0.7187  \scriptsize{$\pm$0.0218}}}&{\small  {0.2593  \scriptsize{$\pm$0.0000}} }& {\small {0.6893  \scriptsize{$\pm$0.0061}} }&{\small {  0.0287  \scriptsize{$\pm$0.0025}}}&{\small{  0.1428  \scriptsize{$\pm$0.0037}} }\\
			\cline{2-9}
			&CDMA&	{\small {0.8602\scriptsize{$\pm$0.0000}} }&{\small{0.1830\scriptsize{$\pm$0.0000}}} &{\small{0.0158\scriptsize{$\pm$0.0000}}} &{\small{0.0171\scriptsize{$\pm$0.0000}}} &{\small{0.4525\scriptsize{$\pm$0.0000}}}&{\small{0.0212\scriptsize{$\pm$0.0000}}} &{\small{0.0093\scriptsize{$\pm$0.0000}}}\\ \cline{2-9}
			&MotifCC&{\small {0.8883 \scriptsize{$\pm$0.0290}} } &{\small  {0.5787 \scriptsize{$\pm$0.1062}} } &{\small {0.7025 \scriptsize{$\pm$0.0653 }} } &{\small {0.0283 \scriptsize{$\pm$0.0087}}} &{\small {0.5029 \scriptsize{$\pm$0.0199} } }&{\small {0.0045 \scriptsize{$\pm$0.0011}} }&{\small {0.0005 \scriptsize{$\pm$0.0004}} }\\
			\cline{2-9}
			&MOGC&{\small {0.9043  \scriptsize{$\pm$0.0000}} }&{\small {0.5841 \scriptsize{$\pm$0.0093}} }&{\small {0.7136 \scriptsize{$\pm$0.0000}}}&{\small {0.4628 \scriptsize{$\pm$0.0010}}}&{\small {0.6988 \scriptsize{$\pm$0.0041}}}&{\small {0.3493 \scriptsize{$\pm$0.0000}}}&{\small {0.1877  \scriptsize{$\pm$0.0000}}}\\
			\hline
			&Motif\_SC& {\small {0.8789  \scriptsize{$\pm$0.0000}}}&{\small {0.4346  \scriptsize{$\pm$0.0000}} }&{\small  {0.4051  \scriptsize{$\pm$0.0000}}}&{\small {0.0400  \scriptsize{$\pm$0.0000}}}&{\small {0.5166  \scriptsize{$\pm$0.0000}}}&{\small  {0.0610  \scriptsize{$\pm$0.0000}}}&{\small {0.0024  \scriptsize{$\pm$0.0000}}}\\
			\cline{2-9}
			&EdMot\_SC&{\small{0.8943 \scriptsize{$\pm$0.0172}} }&{\small {0.3723  \scriptsize{$\pm$0.0933}} }&{\small  {{  0.7050} \scriptsize{$\pm$0.0008}} }&{\small {{  0.3821} \scriptsize{$\pm$0.0132}} }& {\small {{ 0.6136} \scriptsize{$\pm$0.0044}} }&{\small {{ 0.0406} \scriptsize{$\pm$0.0206}}  }&{\small  {{ 0.1710} \scriptsize{$\pm$0.0047}}  }\\
			\cline{2-9}
			${M}_{5}^*$&MWLP&{\small {{ 0.6222} \scriptsize{$\pm$0.0043}} }&{\small {{  0.4324} \scriptsize{$\pm$0.0131}} }&{\small {{  0.1222 } \scriptsize{$\pm$0.0020}} }&{\small  {{ 0.3357 } \scriptsize{$\pm$0.0048}} }&{\small {{ 0.2148  } \scriptsize{$\pm$0.0125}}  }&{\small {{ 0.3349  } \scriptsize{$\pm$0.0013}} }&{\small {{ 0.1532  } \scriptsize{$\pm$0.0038}}}\\
			\cline{2-9}
			&MOSC&{\small {{ 0.8567} \scriptsize{$\pm$ 0.0416}} }&{\small {{0.5124} \scriptsize{$\pm$ 0.0048}} } &{\small {{0.7287} \scriptsize{$\pm$0.0000}} }&{\small  {{ 0.1378 } \scriptsize{$\pm$0.0000}} }&{\small {{0.6692} \scriptsize{$\pm$0.0111}}  }&{\small {{ 0.0316 } \scriptsize{$\pm$  0.0016}} }&{\small {{0.1562 } \scriptsize{$\pm$0.0001}}}\\
			\cline{2-9}
			&CDMA&	{\small {0.8725\scriptsize{$\pm$0.0000}} }&{\small{0.1814\scriptsize{$\pm$0.0000}}} &{\small{0.0830\scriptsize{$\pm$0.0000}}} &{\small{0.0314\scriptsize{$\pm$0.0000}}} &{\small{0.4060\scriptsize{$\pm$0.0000}}}&{\small{0.0153\scriptsize{$\pm$0.0000}}} &{\small{0.0003\scriptsize{$\pm$0.0000}}}\\ \cline{2-9}	&MotifCC&{\small {0.8660 \scriptsize{$\pm$0.0336}} } &{\small  {0.5288 \scriptsize{$\pm$0.0821}} } &{\small {0.7113 \scriptsize{$\pm$0.1615 }} } &{\small {0.0244 \scriptsize{$\pm$0.0042}}} &{\small {0.5356 \scriptsize{$\pm$0.0239} } }&{\small {0.0053 \scriptsize{$\pm$0.0015}} }&{\small {0.0003 \scriptsize{$\pm$0.0002}} }\\
			\cline{2-9}
			&MOGC&{\small {0.9032 \scriptsize{$\pm$0.0000}}}&{\small {0.5815 \scriptsize{$\pm$0.0000}}}&{\small {0.7120 \scriptsize{$\pm$0.0000}}}&{\small {0.3897 \scriptsize{$\pm$0.0000}}}&{\small {0.6032  \scriptsize{$\pm$0.0077}} }&{\small {0.1518 \scriptsize{$\pm$0.0038}} }&{\small {0.1845 \scriptsize{$\pm$ 0.0000}}}\\
			\hline
			&MOGC\_${M}_3^3$\_${M}_3^2$&{\small {{0.9242} \scriptsize{$\pm$0.0127}} }&{\small {{0.5881} \scriptsize{$\pm$  0.0135}} }&{\small {\textbf{0.7373} \scriptsize{$\pm$0.0000}} }&{\small {{0.4717}  \scriptsize{$\pm$0.0000}} }&{\small {\textbf{0.7029}  \scriptsize{$\pm$   0.0061   }} }&{\small {\textbf{0.3878} \scriptsize{$\pm$  0.0100}} }&{\small {0.1764\scriptsize{$\pm$0.0000}} }\\
			\cline{2-9}
			{Multiple}&MOGC\_${M}_3^3$\_${M}_{4}^*$& {\small {0.9134  \scriptsize{$\pm$0.0000}} }&{\small {0.5866 \scriptsize{$\pm$0.0072}} }&{\small {0.7342  \scriptsize{$\pm$0.0000}} }&{\small {0.4416 \scriptsize{$\pm$0.0000}} }&{\small {0.6957 \scriptsize{$\pm$0.0033}} }&{\small {0.3694 \scriptsize{$\pm$0.0000}} }&{\small {0.1914 \scriptsize{$\pm$0.0000}}}\\
			\cline{2-9}
			motif&MOGC\_${M}_3^3$\_${M}_{4}^*$\_${M}_{5}^*$&{\small {0.9193 \scriptsize{$\pm$0.0000}} }&{\small {0.5753 \scriptsize{$\pm$0.0116}} }&{\small {0.7342 \scriptsize{$\pm$0.0000}} }&{\small {0.4325 \scriptsize{$\pm$0.0000}} }&{\small {0.6962 \scriptsize{$\pm$0.0049}} }&{\small {0.3642 \scriptsize{$\pm$0.0000}} }&{\small {0.1718 \scriptsize{$\pm$0.0053}} } \\
			\hline
	\end{tabular}}
\end{table}
\subsubsection{Results of 4-node Motifs and 5-node Motifs}
In this subsection, we show the performances using 4-node and 5-node motifs.
%Because there are 199 4-node directed motifs and 9364 5-node motifs \cite{kashtan2004efficient}, it is impossible to enumerate all of 4-node and 5-node motifs and show their performance.  {\color{blue}On the other hand,  }according to the definition of network motif in \cite{milo2002network}, motif is small subgraph with significant large frequency than that in randomized graph preserving the same degree of nodes, it means not all subgraphs are motif. {\color{blue}Therefore, it is necessary to select the most important one, which encodes the significant patterns in the network. Moreover, mFinder3 \cite{kashtan2004efficient} has shown its effectiveness in estimating the ratio of the number of subgraphs for different motifs by sampling method, it can automatically generate the ranking list of motifs based on a score which is a function of the concentration and the z-score. 
In the experiment, we select the top one 4-node and 5-node motif generated from mFinder3 to test our proposed method. Concretely, according to section 3, we firstly run software mFinder3 to find top one motif, then construct 4-node and 5-node motif-based adjacency matrix according to algorithm 1 in \cite{zhao2019ranking}.  Finally, we conduct comparison experiments on seven datasets using 4-node motif and 5-node motif. For 4-node motif, we show the top one 4-node motif for seven datasets in Table. \ref{motifmfinder} and its corresponding visulization in Fig. \ref{motif}, the top one 4-node motif of six datasets (football, polbooks, Cora, email-Eu-core, Citeseer, Pubmed) are ${M}_4^1$, that of polblogs is ${M}_4^2$. The top one 5-node motif and its corresponding figure for seven datasets are also shown in Table. \ref{motifmfinder} and Fig. \ref{motif}. The results of all higher-order graph clustering methods are shown in Table. \ref{resultRI} and Table. \ref{resultnmi}, MOGC with ${M}_4^*$ or ${M}_5^*$ mean that both edge-based graph and top one 4-node or 5-node based hypergraphs are utilized in Algorithm 1.  

As indicated by Table. \ref{resultRI}, we can know that the performance of MOGC is better than other six higher-order graph clustering methods for almost all datasets, which demonstrates the effectiveness of our method. Moreover, compared to the result of MOGC with ${M}_3^*$, having more nodes in a motif does not lead to better performance. In most cases, the highest accuracy is achieved with ${M}_3^2$, the reason may be the noise in motif with a large number of nodes degrades the performance.  
\subsubsection{Results of Multiple Motifs}
Since our method can utilize the information from lower-order structure and multiple higher-order structures, we conduct experiments to show the performance using edges and multiple motifs simultaneously. It is impossible to enumerate all the combination of all motifs with different number of nodes, so we test the combination of the top one 3-node, 4-node and 5-node motifs shown in Table. \ref{motifmfinder}, the top one 3-node motif is ${M}_3^3$ for all datasets. The results are shown in Table. \ref{resultRI} and Table. \ref{resultnmi}, MOGC\_${M}_3^3$\_${M}_3^2$ indicates that edge, motif ${M}_3^3$ and motif ${M}_3^2$ are utilized in Algorithm \ref{algorithm}, MOGC\_${M}_3^3$\_${M}_4^*$ means that edge, motif ${M}_3^3$ and top-1 4-node motif ${M}_4^*$ are used, MOGC\_${M}_3^3$\_${M}_4^*$\_${M}_5^*$  means that edge, motif ${M}_3^3$, top one 4-node and 5-node motif are used. 

As indicated by Table. \ref{resultRI} and Table. \ref{resultnmi}, compared with other higher-order methods, our method can achieve better results by intergrating multiple higher-order structures. For example, in Citeseer dataset, there are 1255 isolated nodes in ${M}_3^3$ motif-based hypergraph, while there are no isolated nodes in ${M}_3^2$ motif-based hypergraph. The NMI result of MOSC\_${M}_3^3$\_${M}_3^2$ is about 0.13 higher than EdMot\_SC with ${M}_3^3$. As for the hypergraph of motif ${M}_3^2$, even though hypergraph fragmentation does not happen, our method still achieves 0.02 higher than EdMot\_SC in terms of NMI. For Cora dataset, highest rand index result is achieved by MOGC\_${M}_3^3$\_${M}_4^*$\_${M}_5^*$ with edge and multiple motifs. Similar analysis can be made for other datasets. 
\begin{table}
	\vspace{-3cm}
	\caption{Top one of 3-node, 4-node, 5-node motifs generated by mfinder}\label{motifmfinder}
	\centering
	\begin{tabular}{cccccccc}
		\hline
		dataset&football&polbooks&polblogs&Cora&email-Eu-core&Citeseer&Pubmed\\
		\hline
		3-node motif&$M_3^3$&$M_3^3$&$M_3^3$&$M_3^3$&$M_3^3$&$M_3^3$&$M_3^3$\\
		4-node motif&$M_4^1$& $M_4^1$&$M_4^2$&$M_4^1$&$M_4^1$&$M_4^1$&$M_4^1$\\
		5-node motif &$M_5^1$ &$M_5^2$&$M_5^2$&$M_5^3$ &$M_5^2$&$M_5^3$&$M_5^3$\\
		\hline
	\end{tabular}
\end{table}	
		\subsection{Adaptive Node-level Weights Analysis}
\begin{figure}[ht]
			\vspace{-0.5cm}
			\centering
			\includegraphics[width=8cm,height=5cm]{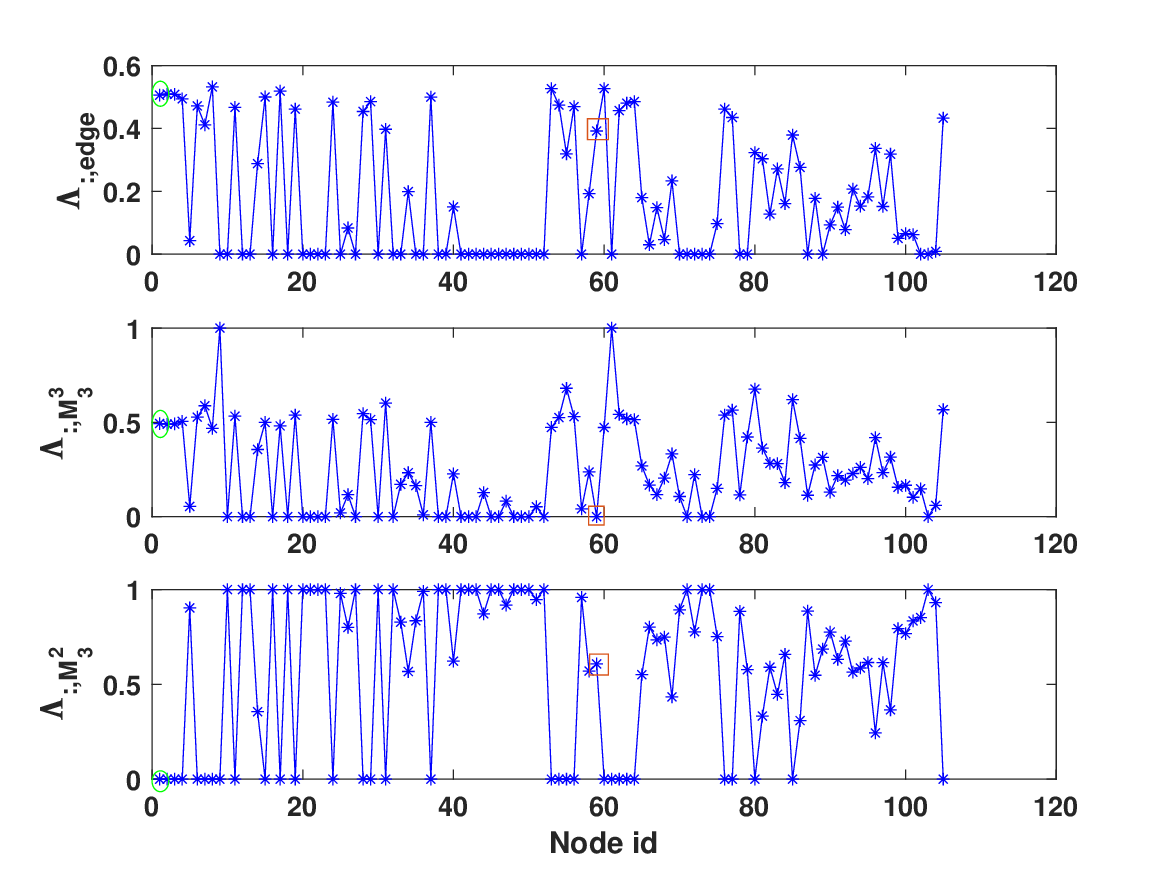}
			\vspace{-0.3cm}
			\caption{The weight $\mathbf{\Lambda}$ of each motif for each node on the polbooks dataset. From top to bottom, three figures correspond to the weight of edge, motif ${M}_3^3$, motif ${M}_3^2$ for each node respectively.}\label{polbooks_lambda}
		\end{figure}

		% our method\_$M_4$\_$M_{13}$ utilizes connectivity pattern from multiple higher-order structures and lower-order structure to enhance the clustering performance. The average weight of each structure ($\mathbf{\Lambda_{:,M_j}}$) learned by proposed method for seven datasets are shown in Table.\ref{ave_weight}. A larger average weight of $\mathbf{\Lambda}_{:,M_j}$ indicates that motif $M_j$ play a more important role in the process of clustering. In addition, SC, Motif\_SC\_$M_4$ and Motif\_SC\_$M_{13}$ apply spectral cluster on different adjacency matrix generated based on different structures (edge/Motif $M_4$/Motif $M_{13}$). We suppose that better performance could be achieved with generated adjacency matrix of structure $M_j$ if $M_j$ play a more important role in the process of clustering. For example, in polbooks dataset, the average weight of $M_{13}$ is higher than others. Moreover, Table.\ref{result1}, \ref{result2} show that Motif\_SC\_$M_{13}$ has better performance than Motif\_SC\_$M_4$ and SC, the clustering results are consistent with average weight $\mathbf{\Lambda_{:,M_j}}$ obtained by proposed method. For other datasets, the clustering results of SC, Motif\_SC\_$M_4$ and Motif\_SC\_$M_{13}$ validate that the calculated weight of each node for each structure by our method is reasonable.  
		The polbooks dataset is chosen to illustrate the performance of adaptive weights $\mathbf{\Lambda}$, the network in polbooks contains 105 nodes grouped into 3 clusters. The 1st-13th nodes belong to 1st cluster, the 14th-62nd nodes belong to 2nd cluster, and the 63th-105th nodes are in 3rd cluster.  Fig. \ref{polbooks_lambda} shows the weights $\mathbf{\Lambda}$ of each motif (including lower-order structure, motif ${M}_3^3$ and motif ${M}_3^2$) for each node, it is apparent that the contribution of structures are different to different nodes. We calculate the average weight for edge, ${M}_3^3$ and ${M}_3^2$, they are 0.1784, 0.2532, and 0.5684 respectively, the average weight of ${M}_3^2$ is higher than others. Moreover, it is shown that Motif\_SC with ${M}_3^2$ has better performance than Motif\_SC with ${M}_3^3$ and SC in  Table.\ref{resultRI} and Table. \ref{resultnmi}, this results validate that the obtained $\mathbf{\Lambda_{:,j}}$ obtained by MOGC is reasonable.
		
\begin{figure}
				\vspace{-3cm}
			\begin{subfigure}{0.5\textwidth}
				\centering
				\includegraphics[width=6cm,height=4cm]{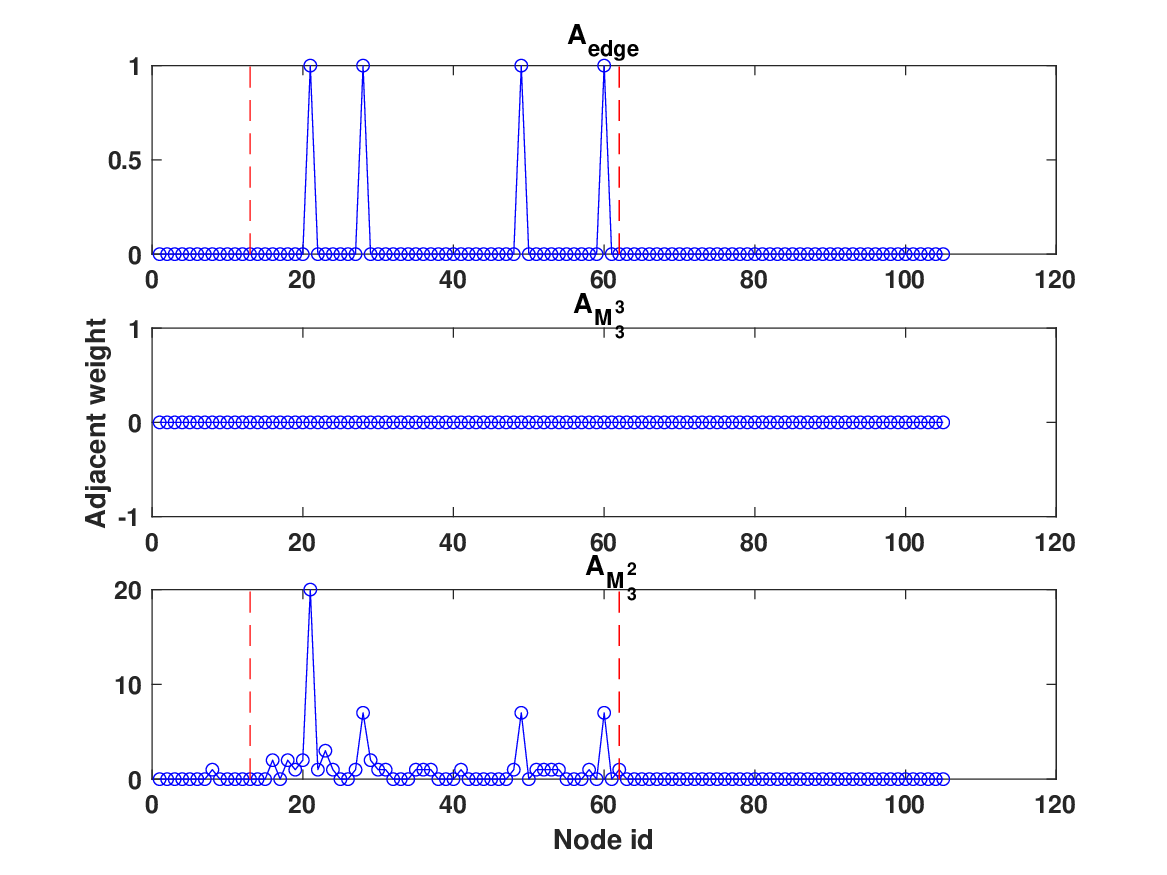}
				\caption{The weight of adjacency matrix for 59th node}
			\end{subfigure}
			\begin{subfigure}{0.5\textwidth}
				\centering
				\includegraphics[width=6cm,height=4cm]{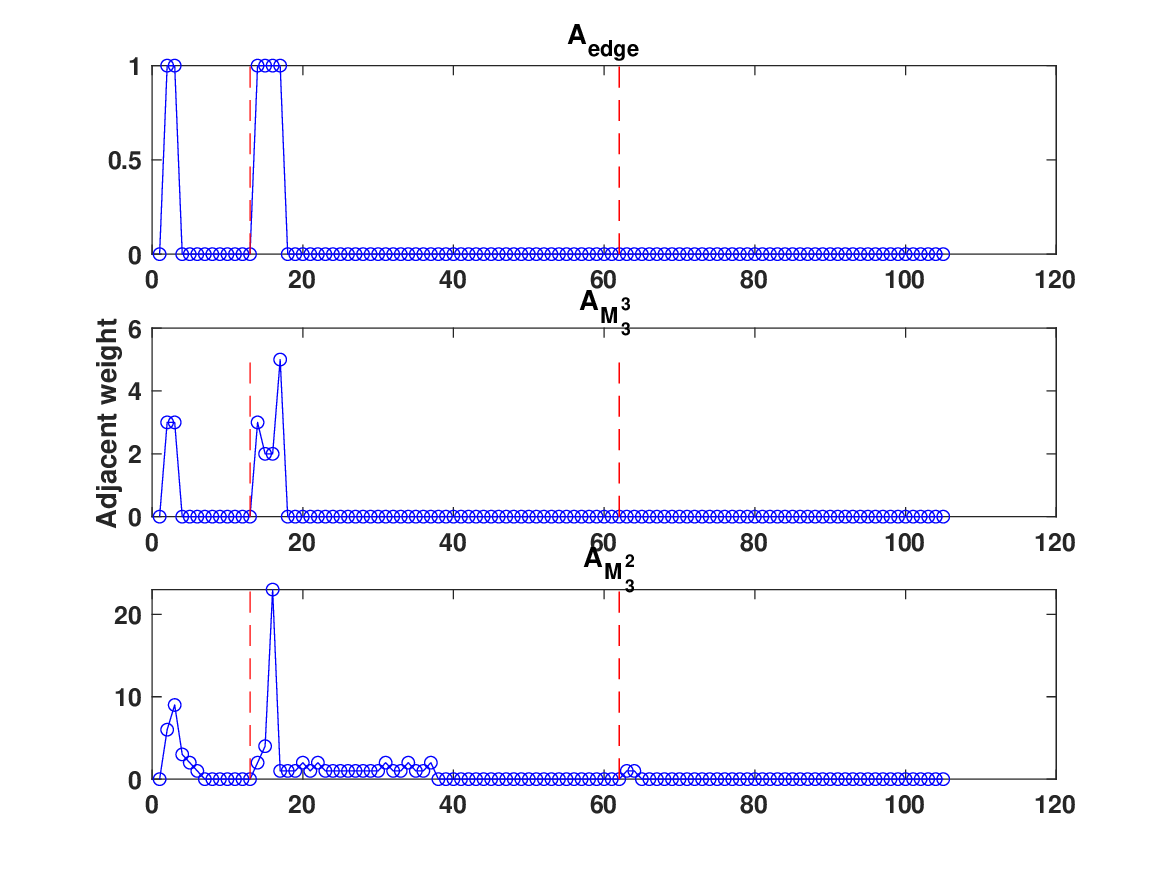}
				\caption{The weight of adjacency matrix for 1st node}
			\end{subfigure}
			\caption{The weight in adjacency matrix for nodes on polbooks dataset. The red dash lines are used to separate the nodes from different clusters.}\label{polbooks_weight}
		\end{figure}

		To further demonstrate the performance of adaptive node-level weight, we checked two typical nodes marked by red square and green cicle. The red square is the 59th node with $\mathbf{\Lambda}_{59,1}=0.3920$, $\mathbf{\Lambda}_{59,2}=0$, $\mathbf{\Lambda}_{59,3}=0.6080$. Fig. \ref{polbooks_weight}(a) shows the corresponding adjacent weight of other nodes connected to 59th node, it is shown that 59th node becomes a isolated node without connectivity with other nodes under motif ${M}_3^3$. 
		From Fig. \ref{polbooks_weight}(a), we can conclude that 59th node belongs to the 2nd cluster, nodes from the same cluster have higher similarity with 59th node than others from different clusters according to $\mathbf{A}_{edge}$ and $\mathbf{A}_{{M}_3^2}$, so both $\mathbf{A}_{edge}$ and $\mathbf{A}_{{M}_3^2}$ contribute to the clustering of the 59th node. However, Fig. \ref{polbooks_weight}(a) also shows that the 59th node has denser within-cluster connectivity with ${M}_3^2$ than that with edge, so $\mathbf{A}_{{M}_3^3}$ can provide more accurate information to the cluster of 59th node. This result indicates that the weight of motif ${M}_3^3$ would be larger than that of edge, which is consistent with calculated $\mathbf{\Lambda}$. The green cicle is the 1st node with $\mathbf{\Lambda}_{1,1}=0.5060$, $\mathbf{\Lambda}_{1,2}=0.4940$, $\mathbf{\Lambda}_{1,3}=0$. Fig. \ref{polbooks_weight} (b) shows the corresponding adjacent weight of other nodes connected to 1st node. The 1st node belongs to first cluster. As can be seen, 1st node has connections with other nodes from 1st cluster and 2nd cluster. Moreover, nodes from 2nd cluster have more connectivity with  larger weights with 1st node under motif ${M}_3^2$, which is not helpful to the correct partition. So it is acceptable that our method assigns 0 to the weight of motif ${M}_3^2$ for 1st node. For lower-order structure and motif ${M}_3^3$, nodes both from 1st cluster and 2nd cluster have connections with 1st node according to $\mathbf{A}_{edge}$ and $\mathbf{A}_{{M}_3^3}$, so it is reasonable that similar weights are assigned to these two motifs.
		%\begin{table}
		%	\caption{The average weight of nodes for each structure calculated by our method\_$M_4$\_$M_{13}$.}\label{ave_weight}
		%	\begin{tabular}{cccc}
			%		\hline
			%		&Edge&Motif\_$M_4$&Motif\_$M_{13}$\\		
			%		\hline
			%		football&\\
			%		polbooks&\\
			%		polblogs&\\
			%		Cora&\\
			%		email-Eu-core&\\
			%		Citeseer&\\
			%		Pubmed&\\
			%		\hline		
			%	\end{tabular}
		%\end{table}
		\subsection{Parameter Analysis}
\begin{figure}[ht]
			\vspace{-0.3cm}
			\centering
			\includegraphics[width=8cm,height=5cm]{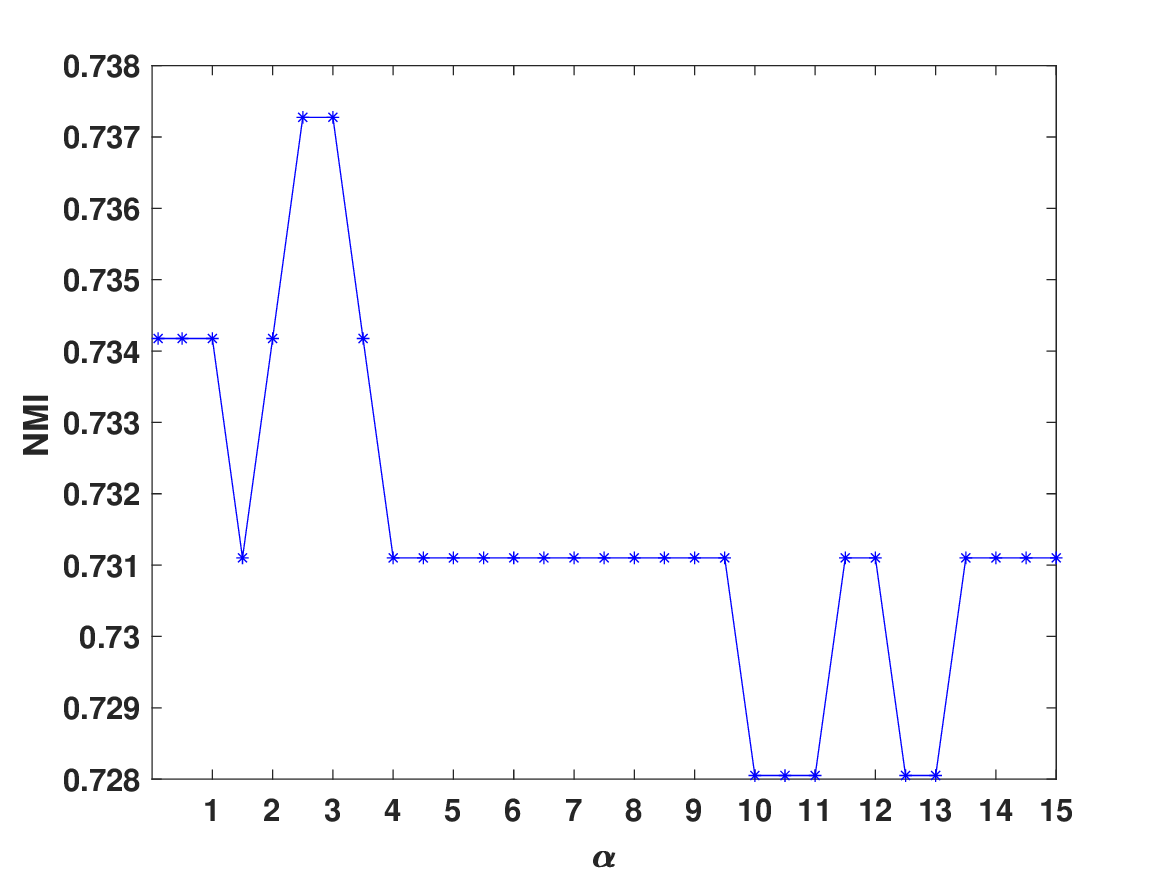}
			\vspace{-0.3cm}
			\caption{Sensitivity of $\alpha$ based on polblogs dataset. Each value is the average value of 20 trials in terms of NMI. }\label{alpha}
		\end{figure}

%		In this section, parameter analysis is conducted to investigate the effect of the parameter $\alpha$ on the performance of MOGC. 
		Theoretically, the parameter $\alpha$ is a trade-off parameter for smooth regularization term, an extremely small $\alpha$ leads to an extremely sparse $\Lambda$ and an extremely large $\alpha$ leads to a $\Lambda$ with almost identical entries. Between two extremas, we obtain suitable solutions. We use polblogs as an example to test.  Fig. \ref{alpha} shows the NMI value of MOGC with edge, ${M}_3^3$ and ${M}_3^2$ with $\alpha$ within the range of [0,15]. It is shown that the optimal $\alpha$ locates within [2,3]. When $\alpha$ is small, i.e., $\alpha=0.1$, the average calculated weights for $\mathbf{\Lambda_{:,edge}}$, $\mathbf{\Lambda}_{:,{M}_3^3}$ and $\mathbf{\Lambda}_{:,{M}_3^2}$ are 0.0969,0.1307, 0.7724. However, when $\alpha$ is larger, i.e., $\alpha=15$, the weight of three motifs are 0.1357, 0.1355, 0.7288, a more smooth solution for $\mathbf{\Lambda}$ would be obtained. This results validates the setting of $\alpha$ should be neither too small nor too large.
		\section{Conclusion}
		In this paper, we propose a multi-order graph clustering model (MOGC) for higher-order graph clustering. Different from the existing methods, the higher-order structures and lower-order structures are integrated at the level of node samples with an adaptive weights learning mechanism. The proposed method can automatically adjust the contributions of different motifs to each node by the adaptive weights. { This approach addresses both the hypergraph fragmentation issue and improves graph clustering accuracy by integrating information from multiple motifs at the node level. }
  Extensive experiments have been conducted to show the effectiveness of proposed method. 
  
  {In the current MOGC model, our aim is to enhance the graph clustering precise by integrating hypergraph structures from multiple motifs linearly. It is interesting to explore the nonlinear fusion of multiple motif-based hypergraphs. As a future research work, we plan to extend the multi-order graph clustering model to incorporate graph neural networks (GNNs). This extension would enable GNNs to achieve superior node representations by integrating higher-order structures from multiple motifs at the node-level nonlinearly. Consequently, this would lead to improved performances in node classification and node clustering.}
  
%		\section{Acknowledgment}
%		This work is supported by Basic and Applied Basic Research Foundation of Guangzhou (2023A04J1682) and the Guangdong Provincial Key Laboratory of Human Digital Twin (2022B1212010004).
		% \begin{figure}[h]
			%	\centering
			%	\includegraphics[width=6cm,height=3cm]{alpha.eps}
			%	\caption{Sensitivity of $\alpha$ based on polblogs dataset. Each value is the average value of 20 trials in terms of NMI. }\label{alpha}
			%\end{figure}



 \bibliographystyle{elsarticle-num} 
 \bibliography{references}





% \end{thebibliography}
\end{document}